STRUCTURE AND INFRASTRUCTURE ENGINEERING
https://doi.org/10.1080/15732479.2024.2311911Taylor & Francis
Taylor & Francis Group

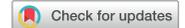

# Asset management, condition monitoring and Digital Twins: damage detection and virtual inspection on a reinforced concrete bridge

Arnulf Hagen[a] and Trond Michael Andersen[b]

[a]SAP Norway, Engineering Centre of Excellence/NTNU, Faculty of Engineering, Trondheim, Norway; [b]Norwegian Public Roads Administration, Division of Operation and Maintenance, Oslo, Norway**ABSTRACT**

In April 2021 Stavå bridge, a main bridge on E6 in Norway, was abruptly closed for traffic. A structural defect had seriously compromised the bridge's structural integrity. The Norwegian Public Roads Administration (NPRA) closed it, made a temporary solution and reopened with severe traffic restrictions. The incident was alerted through what constitutes the bridge's Digital Twin processing data from Internet of Things sensors. The solution was crucial in online and offline diagnostics, the case demonstrating the value of technologies to tackle emerging dangerous situations as well as acting preventively. A critical and rapidly developing damage was detected in time to stop the development, but not in time to avoid the incident altogether. The paper puts risk in a broader perspective for an organization responsible for highway infrastructure. It positions online monitoring and Digital Twins in the context of Risk- and Condition-Based Maintenance. The situation that arose at Stavå bridge, and how it was detected, analysed, and diagnosed during virtual inspection, is described. The case demonstrates how combining physics-based methods with Machine Learning can facilitate damage detection and diagnostics. A summary of lessons learnt, both from technical and organisational perspectives, as well as plans of future work, is presented.**ARTICLE HISTORY**
Received 23 March 2023
Revised 14 November 2023
Accepted 28 November 2023

**KEYWORDS**
Advanced condition monitoring; asset management; bridge monitoring; decision support; digital transformation; Digital Twins; internet of things; predictive maintenance; structural health monitoring; virtual inspection## 1. Introduction

In April 2021 the Stavå bridge, a main bridge connecting the northern and southern parts of Norway, was temporarily closed for traffic. A structural defect had rapidly developed until it had seriously compromised the bridge's structural integrity. A few days after, the Norwegian Public Roads Administration (NPRA), the owner of the bridge, had in place an improvised and temporary solution to the problem. The bridge was reopened with strict vehicle restrictions and 24/7 manual traffic control while a replacement bridge was constructed, a process that, despite executed in a record-breaking speed, would take almost a year.

This paper uses the case to discuss Structural Health Monitoring (SHM) in specific, and asset monitoring in general. It does so in an Asset Management context, building upon and extending work presented in an earlier conference paper by Hagen, Andersen, Reiso and Sletten (2022). SHM is argued as key to achieve Risk-Based Maintenance (RBM), Condition-Based Maintenance (CBM), or predictive maintenance in general. These are methods and principles meant to ensure that maintenance and other in-operation decisions are based on knowledge of the actual, assessed or predicted asset state and criticality rather than on periodic, pre-defined schedules. The present paper also gives more details of the methodological basis and technology used, and reports on recent work on the application of Machine Learning (ML) on the stored data.

Various models for RBM and CBM have been in use for many years. Reliability Centred Maintenance (RCM) has for instance been widely applied in various forms internationally over at least the last four decades, see Emovon et al. (2016) for a review. As a practical example, to be allowed to operate on the Norwegian continental shelf (oil & gas), the maintenance program must be based on a risk-based method, for instance Risk-Based Inspections, or RBI (Andersen & Rasmussen, 1999). The case to be presented shows how online, real-time monitoring such as SHM can be key in supporting methods like RBI. In a situation where the SHM system at the bridge in real time revealed a rapidly increasing risk of failure, a combination of virtual and physical inspections based on the findings made it possible to intervene fast and tackle the situation, and to reduce both the probability and potential consequences of collapse down to what were deemed acceptable levels. The present paper discusses how NPRA, and other road authorities, can use such solutions as a more general tool for bridge inspections, and discusses the challenges a road owner faces when setting out to resiliently implement complex and unfamiliar tools for condition monitoring. It also argues for and demonstrates the value of combining deterministic and physics-

CONTACT Arnulf Hagen ✉ arnulf.hagen@sap.com, arnulf.hagen@ntnu.no 🏢 SAP Norway, Engineering Centre of Excellence/NTNU, Faculty of Engineering, Trondheim, Norway.

© 2024 Informa UK Limited, trading as Taylor & Francis Group

based methods with non-deterministic/probabilistic methods and ML.

The situation under development in April 2021 was detected by remote observation using what constitutes the bridge's Digital Twin (Futai et al, 2022; Hagen, 2017) through SAP Connected Products as shown in Figure 1. Data from Internet of Things (IoT) sensors, accelerometers, and crack/gap gauges, mounted at key locations is sent to a cloud server where it is processed and transformed into physics-based indicators delivering results from, e.g. simulation models or other forms of algorithmic processes. The output can then be processed in ML algorithms that can be deployed to identify complex patterns of behavior and/or detect anomalies. In this context, physics-based approaches are both used as pre-processing before using ML for anomaly detection, and post-processing to improve explainability in cases where the ML models become complex and abstract.[1]

SAP (2023a) defines that a 'virtual sensor—not unlike physical ones—is a device that can be placed at a specific location on [a model of] an item of equipment to provide a continuous reading of the [calculated or measured] physical state of the equipment at this location (for example, with regard to motion, reaction forces, or strain)'.

The principal difference between a virtual and a physical sensor is that readings from the prior result from computer processing, in a sense representing indirect measurements such as from a simulation model or (other) algorithms, while those from the latter come from direct measurement. A virtual sensor may be 'placed' on the same locations as the physical, in that case representing various forms of refinement of the original signal, or on locations where no physical sensors exist, such as might be the case in predictions from simulation models 'driven' by input from control systems or from physical sensors at strategic locations.

In the case at Stavå, the virtual sensors are found at the same locations as the physical ones, but their output—the indicators describing the 'physical state'—is produced through different levels of backend (cloud) processing. It was these indicator values that rapidly increased, alerting the situation. They were also crucial in the online and offline diagnostic work, as well as in the process of localizing the major structural damage behind significant changes in the structural response.

The Polcevera Viaduct in Genoa in 2018 is a tragic example where a known weakness progressed into catastrophic damage, the progression remaining undetected until the collapse. In the aftermath, bridge owners in Italy are obliged by law to follow a sound and structured strategy for bridge monitoring and RBM/CBM. Guidelines issued by the Italian Ministry of Infrastructure and Transport in late 2020 require bridge owners to classify bridges according to its 'attention level' or criticality, and to devise inspection/monitoring regimes accordingly. For the most critical structures, continuous (digital) monitoring is required (Buratti et al., 2021).

Introducing new technology in this sector to solve old (but growing) challenges is also important in terms of recruitment. As competition for skilled labour grows, young graduates often have many alternative options. What does it take for them to choose a career in something as seemingly old-fashioned as road asset management? How can they be convinced to join in to modernise the transportation sector, help it utilize the potential in IoT-based condition monitoring and Digital Twins in an Asset Management context? There is a need to demonstrate the use of advanced technology in conjunction with the societal value of managing road infrastructure in a good way, and to ensure that there is an organization in place for them with a good working environment and exciting tasks. The ongoing generational shift is making this an utterly important issue.

In sum, increased professionalization is required to meet the ever-increasing expectations to more infrastructure to ever-lower cost. There is a need for smarter operation and more targeted maintenance, naturally involving introduction of new technologies and subsequently changed ways of work. However, these are challenging tasks in established organizations with long traditions. Some of these organizational challenges with employing real-time, online monitoring in an asset management context are discussed here. The demand for organizational readiness is also addressed, both on the individual level, the organization itself and the governance systems. A main question is how to arrange internally in organizations like NPRA for all relevant needs, roles, and responsibilities, and to what degree best Asset

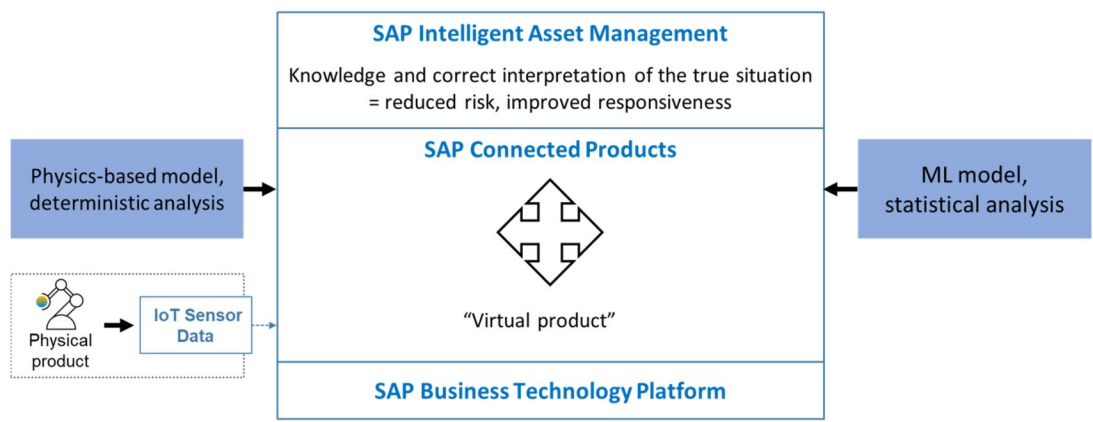

Figure 1. Principal sketch, Connected Products (at time of writing being discontinued) in an asset managment context.

Management practice and principles in other sectors can be reused and adopted for dealing with transportation infrastructure.

A growing interest in sensor-based monitoring as a tool to realize RBM/CBM in the infrastructure sector is manifested in full in the Italian guidelines, having the potential to develop new best practices for others to learn from in the future. Sousa (2018, 2020) presents a comprehensive learning case in this regard, the Leziria bridge project - a long-term project started in 2006 – using a large set of different sensors, and with motivations ranging from diagnosing structural health to enhancing asset management to gaining knowledge for improvement of design guidelines. Butz et al. (2017) on the other hand describes a very specific case using specialised monitoring devices on bridge bearings and expansion joints specifically intended to monitor traffic loads. Along this scale, the present paper aspires to be in the same category of case studies as the Leziria bridge, with the additional plus being that the April 2021 situation demonstrated how technology can be crucial in averting dangerous situations from escalating further. One of the main contributions of this paper is therefore to show the utility and virtue of using online and real time SHM in a practical scenario, and in what turned very fast into a critical situation.

The paper concludes with a summary of lessons learnt, both from an organizational perspective and from the technical side. It also outlines future work needed to improve detection capabilities, in particular through joining non-deterministic and deterministic methods, in this case ML based on probabilistic principles and analyses and simulation based on physical ones.

## 2. Current situation, current challenges

In addition to proving the value in acute situations, the experience to be presented with the Stavå bridge has demonstrated that CBM and SHM partner up well. Through improved methods to detect anomalies, early intervention can be made possible. Referred in Gkoumas et al. (2023, page 7), Tsionis states that 'what is necessary [in SHM] is to link the data to decision parameters', since 'nicely organised data [may not be easy to interpret] by experts to take decisions'. This is an important observation. Improved ability to link data to decisions would potentially trigger corrective measures before an asset's condition has degraded as far as in the case presented in this paper. Experiences from the Stavå bridge project are however that the sound metrics and criteria that Tsionis implicitly refers to do not come out of the box.

There is in general a growing awareness that to keep an aging infrastructure safe in a cost-efficient manner, with condition-based approaches to inspection and maintenance, online and real time structural monitoring plays a key role. The European Commission has launched several initiatives in this respect (see for instance Gkoumas et al., 2023, and COST Action 1402 in Structural Health Monitoring). Germany, as one of the largest countries in Europe when it comes to transport infrastructure, has put the subject high on the agenda with a dedicated digitalisation strategy for the transport sector, and a combined ministry of transport and digitalization (Bundesregierung, 2022). And in Italy, the implementation of the mentioned new guidelines is well underway, with a significant number of bridges already classified and the process of fitting bridges with the required physical online monitoring infrastructure started.

### 2.1. The Norwegian public roads administration (NPRA)

NPRA is responsible for the construction, operation, and maintenance of the main road system in Norway. This includes approximately 10 600 km roads, 580 tunnels, 42 ferries within 16 routes and 5 800 bridges. There is a large ongoing improvement program in NPRA within Asset Management with the aim to improve uptime and road safety, and to reduce environmental and climate impact, while keeping costs from escalating, all as instructed by the Norwegian Ministry of Transport. Asset Management is in turn inspired by the International Standardisation Organization (ISO) 55000 standard (ISO, 2014) and by so-called best practices in national and international industry. Projects prioritized the following years include to:

- implement software for Asset Management (including new work processes, competence, and organizational adjustments),
- purchase and implement software for technical documentation (design, construction, and operation),
- test and implement operational support centres/functional support, and
- develop and implement new strategies for long term planning and operational decision support based on knowledge of risk and technical condition, as well as on the principles of ISO 55000, RCM, and other best practice tools within Asset Management.

R&D activities are carried out together with research institutions and other partners to test models for condition monitoring of bridges, roads, tunnels, as well as the road surroundings (hazards such as avalanches, landslides, and drainage). Also, different climate-friendly solutions are tried out and partially implemented, such as electric and hydrogen-based ferries and vehicles for winter (e.g. electric snowploughs) and summer operations. The use of Building Information Modelling (BIM), Digital Twins and ML are also demonstrated in various applications, as are uses of SHM. While there is extensive research in this field, there is a large gap between the knowledge and experience gained in academia and R&D institutions, and its adoption in the industry. See for instance Cawley (2018) for a comprehensive discussion on the need for such real-life implementations. The Stavå bridge project is through his 'check list' (ibid. page 1241) well placed as a favourable candidate for early adoption of SHM:

- No over-designed, built-in reliability. When the bridge was designed and built in 1942 usage and operational

conditions were obviously entirely different, with slower speeds, lighter vehicles and less traffic.
- Severe economic and societal (e.g. preparedness during crises) consequences of structural failure. The bridge is a main connection between north and south of Norway, and alternative paths for trucks are long.
- Critical damage occurs in short time. The reason why the bridge was originally selected for SHM monitoring (pilot in 2016, full campaign in 2019) was the suspicion that structural failure could develop fast, a suspicion the experience from 2021 confirmed in full.
- Current inspection methods are inadequate. Manual inspection of the bridge was costly, with need for single lane crossing and traffic control on both sides. It also interfered with the traffic and made it impossible to get a correct picture of the dynamic behaviour of the bridge, as vehicles slowed down and changed driving paths during the inspection.
- Regulations/standards govern the process. Regulations require the capacity of structures like bridges to be no more than 100% utilised, or always have reserve capacity, according to standard code (though it is not apparent in what way this code should be applied with respect to findings from SHM).[2]

One key SHM project in NPRA has been the Stavå bridge. It has not only been of crucial importance to ensure safe operation of the bridge itself, but also a valuable learning case. It has given insights into organizational challenges that need to be tackled to release the potential gains of Internet of Things (IoT) and Digital Twins, in addition to relevant requirements to software for collection, storage, and analysis of data. It has also become apparent that third-party expertise in analytical work and providing input to decision making is required, be that technical consultants, universities such as the Norwegian University of Science and Technology (NTNU), or R&D institutions like SINTEF.

## 2.2. New technologies equals organizational challenges

When assessing whether to implement new technology, the TRL-level (Technology Readiness Level) is crucial. What is the level of technology maturity, and are the services well-proven and guaranteed for? Can such services be procured and taking directly into practical use? The NPRA organization does not itself perform operation, maintenance, and construction of its highways, but rather purchases such services in the market. It is paramount that the suppliers can be trusted to deliver and that their products are robust and reliable. Along this scale, SHM based on accelerometers as at Stavå is at TRL 7 (prototypes tested in operational environments), also as concluded in Gkoumas et al. (2019).

A less used success criteria is Organizational Readiness to Change (ORC), meaning how people, processes and systems are prepared to embrace new ways of working, and new tools to support them. See e.g. Robertson (2021) for a comprehensive review. ORC is on the agenda in NPRA, as without sufficient attention to the degree to which the organization can adapt and change, it will be difficult to implement new technologies, methods and processes. The end goal is to improve in a resilient way, such that the improvement is robust when it comes to changes in key personnel, management, software architecture and market. This is of particular importance with advanced services like SHM of bridges and other complex structures or systems.

The way NPRA assesses its organizational readiness level is currently through rudimentary checklists of issues that must be considered from an idea/project is conceived until it is resiliently implemented. For the utilisation in NPRA of SHM in general and Digital Twins in specific, to support the objective of implementing RBM/CBM, this means to consider e.g.:

- Personnel competence and capacity: Does the organization have the people needed today? What will be the requirements and plans to build up the necessary capacity and competence?
- Organization, roles and responsibilities: Are there defined roles in the organization for this today? What is the correct balance to strike between internal capacity and external support (services, research)? How can an internal expert group be built up that focuses on condition monitoring using IoT sensors on state-owned critical bridges and other structures?
- Work processes: Are work processes describing best practice in place today? What must be developed because of more advanced condition monitoring? When a situation develops fast and urgent action is needed, such as in April 2021, who is in the frontline to process and assess warnings? Who do you call, even during nighttime, and what is their mandate? And who makes sure the monitoring system is up and running?
- Rules and regulations: Are there existing regulations that limit the potential of condition monitoring today? Will mandatory checklists based on physical inspection continue to take precedence over discoveries from virtual inspections of SHM systems? Must codes, rules and regulations be changed, and if so, how?
- Metrics and decision criteria: Among the first measurements a medical doctor makes on a patient is to read temperature and measure blood pressure. They are well-established metrics, main indicators of human health and with clear criteria distinguishing normal from not normal, key to e.g. telemedicine. Which metrics should be used to measure structural health during remote monitoring, or 'tele-engineering'? And what decision criteria can decision makers resort to when evaluating whether or not to trigger potentially costly action?
- Contracts/purchasing: Does the organization know how to purchase these services today? Are the procurement processes fit for purpose? Are adequate competencies (and products) available in the market? Is there a need for new kinds of frame agreements?
- Security and risk: The national highways, particularly the bridges, tunnels and similar structures, are critical parts of national preparedness (hospital, firefighting, police, military use). They are important to protect against sabotage in various ways (physical, people or digital). The use of advanced condition monitoring can be an important tool to improve resilience, but can it also be a target for actors with malign intent?[3]

### 2.3. Reliability- and condition-based maintenance (RCM/CBM) in NPRA

NPRA has successfully tested RCM in several pilots. The revised strategy for operation and maintenance will be based on such methods in a broad risk perspective, looking at the effects of failures and maintenance work on the entire transport system. The importance of this has gained significant recent attention, particularly as climate changes wreak havoc on old truths. NPRA (2022) states the need to implement periodic vulnerability assessments of the road network, and to give special attention to detected vulnerabilities. Bridges are particularly mentioned as assets where increased frequency and severity of flooding and erosion have increased beyond the original design assumptions, and thus increases risk of failure. This may in turn be reason in and by itself to change the inspection regime from the original one, in effect introducing RBI as a tool.

The main objectives are in general to do the right maintenance, at the right time and in the right way, and by this be able to implement predictive maintenance and plan and schedule well ahead of time. CBM is a preferred strategy when there exists a reliable and cost-effective method using new technology, reliable sensors and good cloud solutions for storage and analysis of large volumes of data. Information of technical condition is used for (Norwegian Ministry of Transport, 2021):

- Operational decision support (days, weeks, and months).
- Tactical decision support (the next budget years).
- Strategic decision support (5–10 years planning).

The technical condition of assets combined with the associated risks form the basis to plan and prioritize maintenance or renewal (investments). The current state in NPRA is that the pavement condition is monitored on a yearly basis, the road fundament is inspected with laser/geo-radar on a regular basis, while bridges are inspected and classified every year/every five years, as are tunnels and road surroundings. The Norwegian Ministry of Transportation has an ambition to establish revised methods and indices for assessing and maintaining the technical condition of the various road objects, including the county roads (a total of 44 000 km). The mentioned process of adapting to climate change is an example of this. NPRA is central in the work, in cooperation with other road owners in Norway.

### 2.4. Asset management for road infrastructure

Public institutions, private companies and other organizations that work with road transport infrastructure are found in all countries. Being the nerve pathways for land-based private and commercial mobility, it is one of the world's largest sectors in terms of value and employment (World Economic Forum, 2016) and is fundamental for efficient operation and economic development of a region or a nation. To ensure good mobility, there is a need for a robust road transport system. This includes not only development of new infrastructure, but also improvement, operation, and maintenance of the existing one in the best possible way. For a long time, efforts have been made to develop methods and tools that can be used to optimize this process, all under the heading Asset Management.

Asset Management is a huge field internationally. It covers philosophies, principles, methods, and tools for dealing with critical infrastructure such as buildings, factories and production facilities, all forms of transport infrastructure (road, railway, maritime, air, aerospace), and energy production on land and at sea. Different industries and sectors have each developed practices based on their specific needs and frame conditions and have to some extent assigned different names and contents to the methods for Asset Management. Examples of components within this frame are, in addition to RCM, RBI and CBM: Total Productive Management, Six Sigma, Lean, Concurrent Engineering, Dependability or Reliability Management. (For reference, see Diaz-Reza et al., 2022; Emovon et al., 2016; IEC, 2014; Pereira Cabrita et al., 2014)

The drivers for improvement have often been competition and business needs (quality, production efficiency, costs), or safety (risk of accidents, emergency preparedness, new requirements). Compared to some other sectors, one can argue that the Asset Management practice is generally less advanced in applications for road infrastructure, reasons possibly to be found in a relatively low degree of internationalization and standardization, as well as fewer companies offering such services internationally/cross-border. Uncertainty about cost vs. benefit of using methods like remote monitoring plays however an important role, as an SHM system normally is a significant investment and as it often is difficult to quantify the value of more precise knowledge of the true state of the structure. See Sousa et al. (2022) or Bertola et al. (2020) for comprehensive discussions on ways and use cases attempting to quantify the Value of Information (VoI) that can result from SHM, and how this can be used in decisions on whether or not to invest in it.

This general lack of a universal common ground means more local and national solutions for how to organize, finance, and manage the road sector, leading to:

- Different priorities within each country.
- High degree of nationally developed IT tools.
- Mostly nationally based companies providing services in the field of construction, operation, and maintenance.
- National standards and regulations.

This applies especially to operation and maintenance and smaller construction projects, while there for larger investment projects is a greater degree of internationalization, with multinational consultancies, construction firms and technology providers offering services across borders. There are, of course, exceptions. Transnational laws and regulations exist that impose requirements regarding the management of road infrastructure within Europe, e.g. Tunnel Safety Directive (European Commission, 2004) and across US states, e.g. Asset Management principles (US Department of Transportation, 2007). Other characteristics of road transport infrastructure, its management and context are:

- Open system. This is quite rare, and the opposite of e.g. railways, which might help explain why Asset Management historically has been more popular in companies managing railway infrastructure than road infrastructure.
- Close dependency between Asset Management (creation, operation, maintenance and renewal) and Transportation (usage). This affects criticality, risk and performance requirements, and is dynamic because the usage, requirements and even climatic conditions change over time. Hence, the term 'Value Driven Asset Management' is relevant, using more money on maintenance if the safety or uptime is improved (optimizing value of transportation vs. cost).
- Large variations between countries. Different countries have different frame conditions, such as population sizes, transport distances, ways of integrating with other forms of transport providers, large cities vs dispersed population, climate, geography, and topography.

### 2.5. Professionalization in the road infrastructure sector

There has in general been limited interest in investing sufficiently in professionalization of Asset Management by road owners. Businesses follow the money, and there have been larger investments in Asset Management for other sectors like the oil, gas & energy sector, medical & pharmaceutics, production industry and railways. This is however changing, with important trends affecting road owners arguably being:[4]

- A more automated, digital, and interconnected transport system is emerging, the effect of which is a more efficient, but also a more vulnerable road system. This places new demands on Asset Management pertaining to organizational issues, personnel competence, capacity levels, procurement, as well as contract formats, work processes, methods, and tools for Asset Management.
- New technologies and new methods provide new opportunities and new challenges for monitoring of infrastructure and in measuring the condition of various infrastructure elements. ML has proved to be very promising in analyzing large amounts of data for decision support, examples including automatic measurement of pavement condition, road bearing capacity, road marking condition and guard railing condition. (Ranyall, et al. 2022; Tomasgard 2023) Fiberoptic sensors may detect incidents and record traffic in tunnels (Monsberger et al. 2018), drones can automatically inspect bridges and other infrastructure (Ebbesen 2022), while data from regular vehicles can help assess road condition (Arvidsson et al. 2022). By combining multiple data sources, a more comprehensive basis for planning and prioritization can be established.
- Demands from the public are ever-increasing regarding safety, efficiency, and increasingly environmental impacts. Users expect to be immediately informed about the status of roads and planned road work, and to be advised on transport routes and notifications along the driven route. This translates to a need for constant development and improvement in Asset Management for road infrastructure.
- Automated transport technology for cars, commercial transport and special vehicles for operation and maintenance is in rapid development, imposing new requirements on infrastructure, design, operation, and maintenance. An example is automated, position-based snow removal, salting, and sanding. Another is autonomous ferries across fjords, eliminating the need for a captain and a crew. But while 'human-related errors may be reduced [ … ] new safety issues related to, for example, the reliability of the technology and cybersecurity, arise'. (Guo et al., 2021)
- Fossil-free transport and electric vehicles create the need for electric power along roads. An example is testing electric pantographs for heavy transport in Sweden (Ames, 2016). When most trailers in the future expectedly will be powered by electricity, charging facilities must be established, operated, and maintained.
- Increasingly multimodal logistics require better integration between transport modalities, road, rail, sea, and air, resulting in increased complexity and demand for seamless user experience and technical integration.
- Transport infrastructure is increasingly critical to society and of strategic importance to areas such as military defence and emergency preparedness (health, fire, police, rescue), and security pushes up on the agenda. There is a need for vulnerability analyses and understanding of risk to reduce it to an acceptable level and to keep the road systems resilient and secure. Physical and organizational measures are required, as well as practice in how to handle critical events. Wars and heightened levels of tension impacts Asset Management. How well protected are the technical rooms that contain control electronics for tunnels? Or the access to technical rooms on both sides of a highly congested bridge? How do we determine when 'secure' is 'secure enough'?
- Data security, data anonymization and digitalization is increasing in importance, as the prevalence of digitalization in all aspects of society introduces new challenges and threats. Privacy regulations like EUs General Data Protection Regulation (GDPR) must be adhered to, even more so as image recognition becomes prevalent, and large amounts of data must be selectively stored, efficiently analysed, and properly used. The introduction of new IT tools requires change and development of

- employee knowledge, new organizational roles and responsibilities, and a practice that ensures continuous improvement.
- Climate change means hotter, wetter, dryer, wilder and more unpredictable weather. Ongoing changes in the climate will have an impact on the design, construction, management, operation, and maintenance of road infrastructure, affecting various countries and various parts of the same country widely different.
- Critical accidents/fires in tunnels and bridge collapses make Asset Management increasingly topical, creating demand for better practice and introduction of new laws and regulations. Such incidents will continue to be driving forces for change and improvement also in the future, especially in societies with diminishing tolerance for personal risk.
- Further professionalization of Asset Management is required from and by road owner organizations. Coordination across continents and countries will continue, and the implementation of global standards will likely increase (presuming globalization prevails).
- Better and more internationally oriented IT support tools are needed for Asset Management, and the supply is growing. Greater interest in road infrastructure among suppliers who have traditionally been involved in other markets leads to further expansion of businesses offering more standardized services and tools for road owners internationally is expected.
- There is an increasing focus on innovation, development, and improvement work. Use of more agile models for development of IT functionality and related processes is expected, as is closer collaboration between road owners, commercial companies, universities, and research institutions. Models for innovation partnerships and an increased degree of innovation built into service and maintenance contracts will be more common.

During work with an internal report yet to be published, NPRA calculated the overall costs for the incidents at Stavå bridge and Badderen bridge, another key bridge on E6, this one connecting two regions in northern Norway (Amundsen, 2022). The calculations considered both the direct costs to NPRA, as well as costs to users and to society related to delays and longer transport routes. In the case of Badderen the alternative route from one side of the bridge to the other passed through Finland and was more than 600 kilometres long. The report concludes that both these incidents might have been avoided, or at the very least become less severe, with proper and systematic use of condition monitoring and predictive maintenance. In other words, by having acted before the condition got so poor as to limit traffic, and with enough time to ensure a best possible and properly planned solution for renewal or repair, the economic and societal consequences would have been significantly less. For each of these two incidents, the internal analyses show that more than 10 million Euros could have been saved in direct and indirect costs with earlier action.

In a more strategic perspective, the value for engineers when developing new bridge designs should not be underestimated. This was the motivation behind a newly started project where SHM is to be used on the timber bridge Norsenga (Arminas, 2023; Eiterjord, 2023). The intention is to better be able to understand the behaviour in operation of these kind of bridges with novel designs, of which one collapsed dramatically and entirely unexpected in 2022, barely ten years old (Barker, 2022; Bridge Design and Engineering, 2022).

### 2.6. SAP[5] and Connected Products[6]

Key to CBM, RCM and similar methods is understanding the state and condition of the system in question. The better insight, the better chance of assessing the risk, both understanding when a situation requires action to be taken and in part also what may be the consequences of not acting. Online and continuous regular IoT sensor data feed is an enabler for the application of predictive maintenance solutions; the more is understood of the current state and behavioural history of an asset, the more precise assessments can be made regarding its expected behaviour in the future. Two principally different methods of doing such assessments are analysis based on first principles, e.g. natural law (as encoded in algorithms and simulation models), and analysis based on the law of large numbers, comparing a situation to what is expected (as encoded in ML). Far from being exclusive, these have a symbiotic relationship, complementing each other. One does not need machine learning to re-invent Newton's Laws of Motion, but one may need it to detect inconsistencies and apparent deviations from those laws in the concurrent readings among a large mix of single- and multi-axes sensors placed across the asset.

In any case, to provide real and precise insight, sensor signals may need to be refined to remove for instance bias and noise, and to describe the asset state as closely as possible in terms familiar to the user (Hagen, 2019). SAPs Connected Products (CP) deals with this, giving managers and engineers tools needed to understand the performance of assets in operation, and to analyse such operational data and use it for qualified maintenance decisions. CP offers a platform for deployment and operation of sets of continuously running applications or models (Digital Twins), like in an SHM setting, and connect them to the IoT sensor data that streams from the physical assets. The objective is to encode physical principles in algorithms and simulation models, use them to refine data, and thus make both anomaly detection and diagnosis more transparent. Users can develop and deploy such general or domain specific algorithms that will sift through the incoming IoT sensor data, filter, refine and align it, and in combination with other means (such as ML tools) detect and notify anomalies. Based on notifications from the system, the user/expert can then investigate details through traditional plots or 3D visuals in the user interface, embedding complex finite element models for the specialists (while striving for simplification for non-expert responders and decision makers), and take actions accordingly.

The general flow is described in Figure 2. A sensor signal is sent to a cloud server, where it is buffered in a data lake and if needed serialized and temporarily stored in a CP database. Then it will, either in batches or continuously (in practice in micro batches as short as seconds), be pre-processed before being subject to a sequence of analyses, either through ordinary algorithms or through (traditional or inverse) simulation models. The result may be presented as readings from virtual sensors. In the process, 'chunks' of high-resolution data of possible interest are archived according to user defined retention rules, while the remaining is deleted to avoid unnecessary storage. The solution is sensor agnostic, designed to consume most types of time series from sensors (or APIs). The output from the virtual sensors is subject to further analysis, using traditional threshold-based techniques or more advanced ML methods, and presented to the user for decision making or further analysis.

## 3. The Stavå bridge project

The 100-meter long Stavå Bridge as shown on Figure 3 is an arc bridge built in reinforced concrete more than 80 years ago. It has over the decades experienced gradually increased traffic, particularly from heavy vehicles, resulting in loads far exceeding those for which the bridge was initially designed and constructed. The increased speeds and weights, or traffic load in general, have over decades caused intensifying wear and tear on the bridge. Consequently, it has for the last decade or so been subjected to increased scrutiny in the form of regular and frequent physical inspections.

Nonetheless, quite often a change in behaviour is in the dynamic response, as depicted in Figure 4, affecting how a structure responds under varying load conditions, altered traffic situations, environmental changes or other influences. This is particularly so here, with a bridge designed at a time when traffic was both significantly slower and significantly lighter. Thus, while traditional campaign-based inspections involving traffic control and closed-off lanes successfully have detected and monitored the development of cracks at rest, scaling, and other visible damages, it has not given equally good insight into the bridge's response to the dynamic loads.

For instance, measuring static crack width in reinforced concrete structures tells only part of the story. The behaviour of the cracks during traffic may be needed to tell the rest. The max-min range of the crack gap opening during the crossing of a truck, such as shown in Figure 5, may at times be more important than measuring the gap in undisturbed condition. A Fast Fourier Transform (FFT) as seen in the right graph in the figure shows a low frequency component proportional to the speed of the vehicle (qualified by vehicle length, the period of an opening-close cycle). The higher frequency components of the signal may on the other hand reveal mechanisms driving crack growth. In the case shown in the figure, the high frequency component of about 1.4 hertz (Hz) matches the frequency of the first (lateral) natural mode of the bridge, indicating that sideways motion may be correlated with crack development, aiding further diagnostics.

In general, while physical inspection can reveal defects visible on the outside, such as cracks, internal defects may in the early stages only be observable as changed motion characteristics, alternatively through elaborate and expensive (and likely disruptive to traffic) Non-Destructive Testing (NDT) like ultrasound or ground penetration radar. Weakening and stiffness reduction from internal slippage or corrosion of rebars may be discovered through local or global change in dynamic behaviour before it can be observed as static changes or as damages detected in a visual inspection. Moreover, as the Stavå bridge is non-symmetric over all axes and worn and torn over the lifetime, one vehicle crossing might cause an entirely different response than a similar vehicle passing at a slightly different speed, or on a slightly different path, while northbound traffic naturally creates other responses than southbound, due to sloping and differences in curvature in the north and south halves. These factors make the structural assessment quite complicated, and the bridge was therefore fitted with a larger number of sensors than might have been the case in a simpler structure.

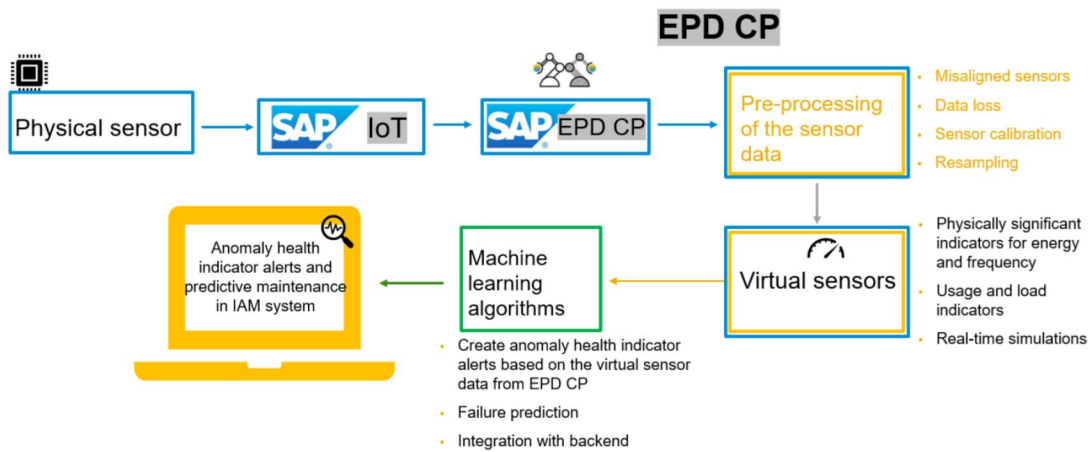

Figure 2. Schematic flow from sensor on a physical asset to the end user (human or machine).

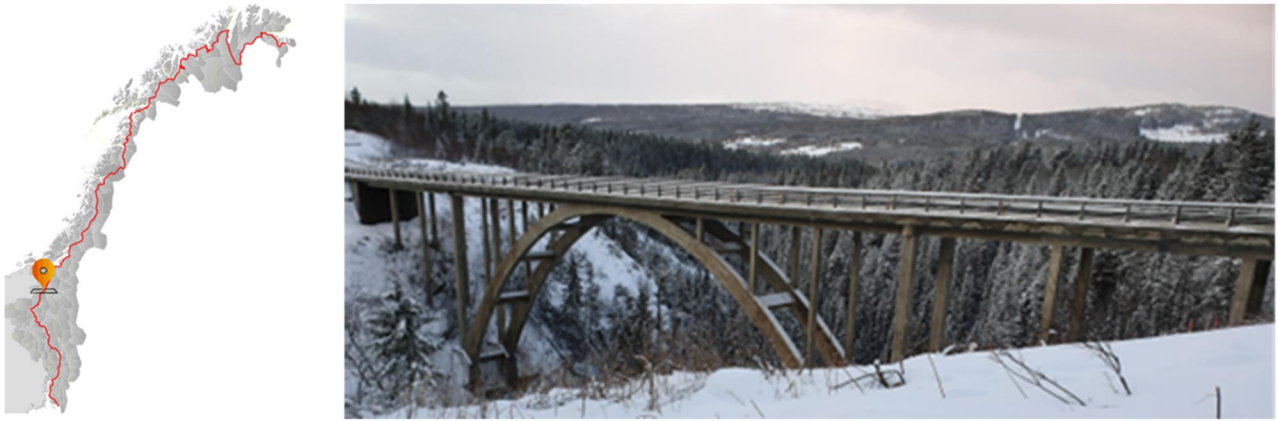

Figure 3. Stavå bridge on E6 as seen from the east, trondheim to the left, Oslo to the left, tying together northern and southern Norway.

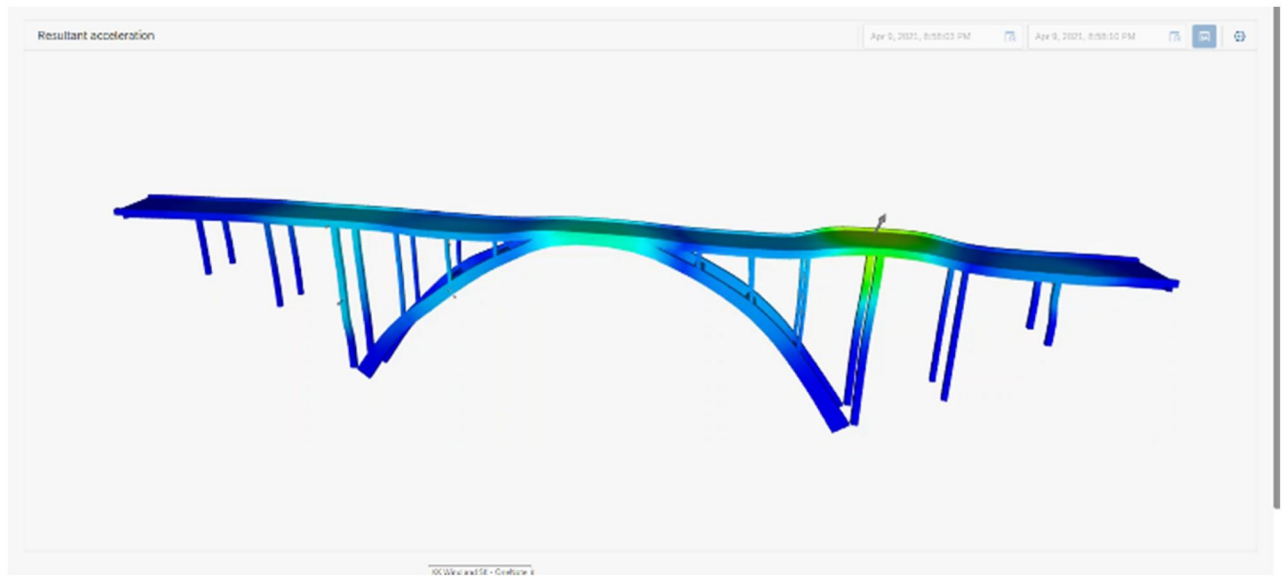

Figure 4. User interface example, here a screengrab of situation playback showing acceleration vectors.

### 3.1. Methodology and configuration

Deciding on numbers, types, frequencies, and positions, requires general domain- and system/structural knowledge, depending on the objective and ambition of the monitoring effort. The sensor types and their positions shown in Figure 6 were chosen by domain experts and bridge inspectors building upon experiences from an initial monitoring phase:

- 21 wireless and battery-driven[7] tri-axial accelerometers (B-M, O-X), sampling at rate 64 Hz, to gain sufficiently complete understanding of the bridge's global behaviour.
- Four wireless, battery-driven gap sensors (I1-2, S1-2), sampling with rate 128 Hz, fitted on cracks located symmetrically on both sides of the road centerline about five meters north and south of bridge span midpoint, to monitor development of static, quasi-static and dynamic behavior.
- Four wireless, grid-connected distance gauges (A1-2, N1-2), sampling with rate 128 Hz, mounted on each side of the expansion joints at both abutments.

A first installation in 2016 consisted of six accelerometers mounted on the guard rails. Two were placed opposite each other on the bridge deck at section 4.4 as shown in Figure 6, the four others placed on the west side at 1.0 (some meters south of north abutment), as well as west on sections 4.2, 4.6 and 5.0. After a major event in November 2018, described in the next chapter, the installation was significantly extended in 2019. Throughout the paper, sensor positions will where relevant reference Figure 6 in either or both ways:

- Through lettering. B to M to the right when driving southward, O to X to the left, except S located in the middle of a transverse beam at section 4.0).
- Through a combination of section, side and elevation: From 1.0 at the northern abutment to 8.0 at the southern, location H to the right when driving southward and V to the left, and index 1 (like in V1) signifying location under the bridge deck, 2 on a pilar/column, and 3 on foundation (4.0).

The orientation of axes of the bridge's global coordinate system, and rotation about axes, follow the right hand rule:

- x-axis: Longitudinal, parallell with main bridge span (straight segment between sections 4.0 and 5.0), positive towards south.

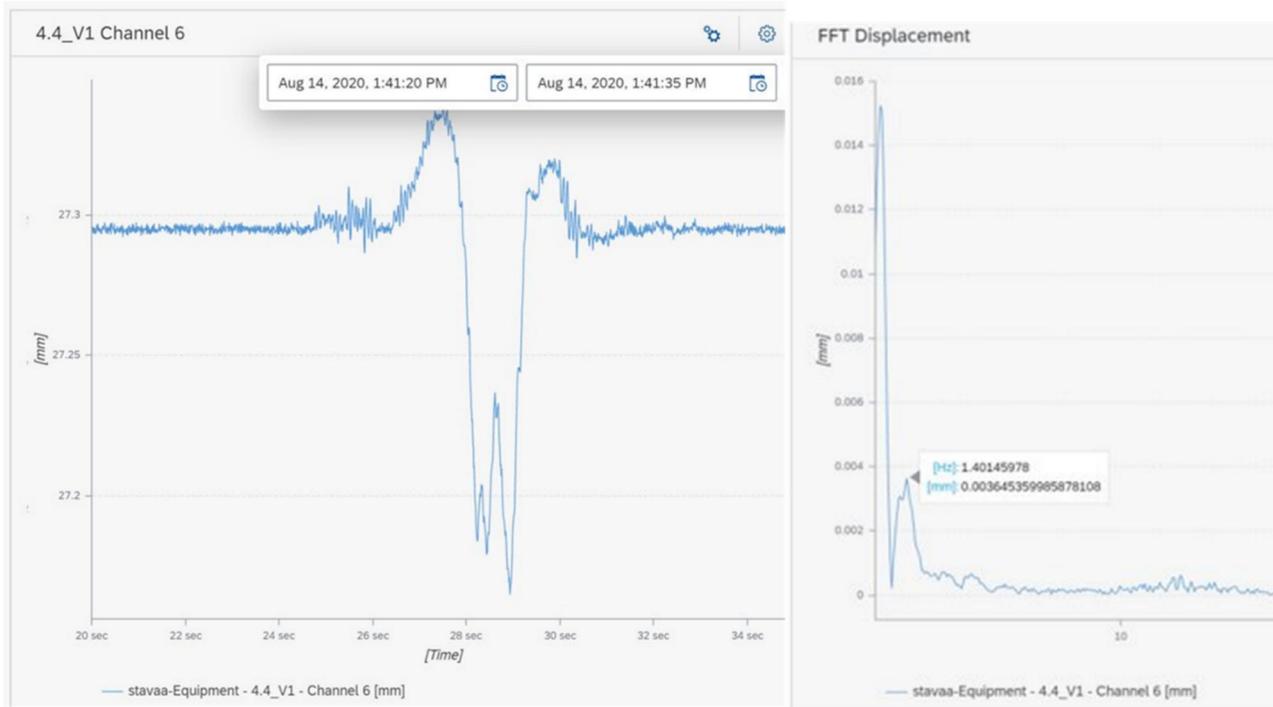

**Figure 5.** Dynamic response in crack gauge S1 in Figure 6, upon passage of a trailer truck (two mass centroids, as seen by the hump in the middle of the signal in second 28–29). The left graph shows the time domain view of non-calibrated distance gauge length, millimetre along the y-axis and time in seconds along x-axis (note that smaller measured value on the gauge equals widening of crack opening). The right graph plots the same signal in the frequency domain using FFT, normalized amplitude along y-axis, frequency in Hz along x-axis.

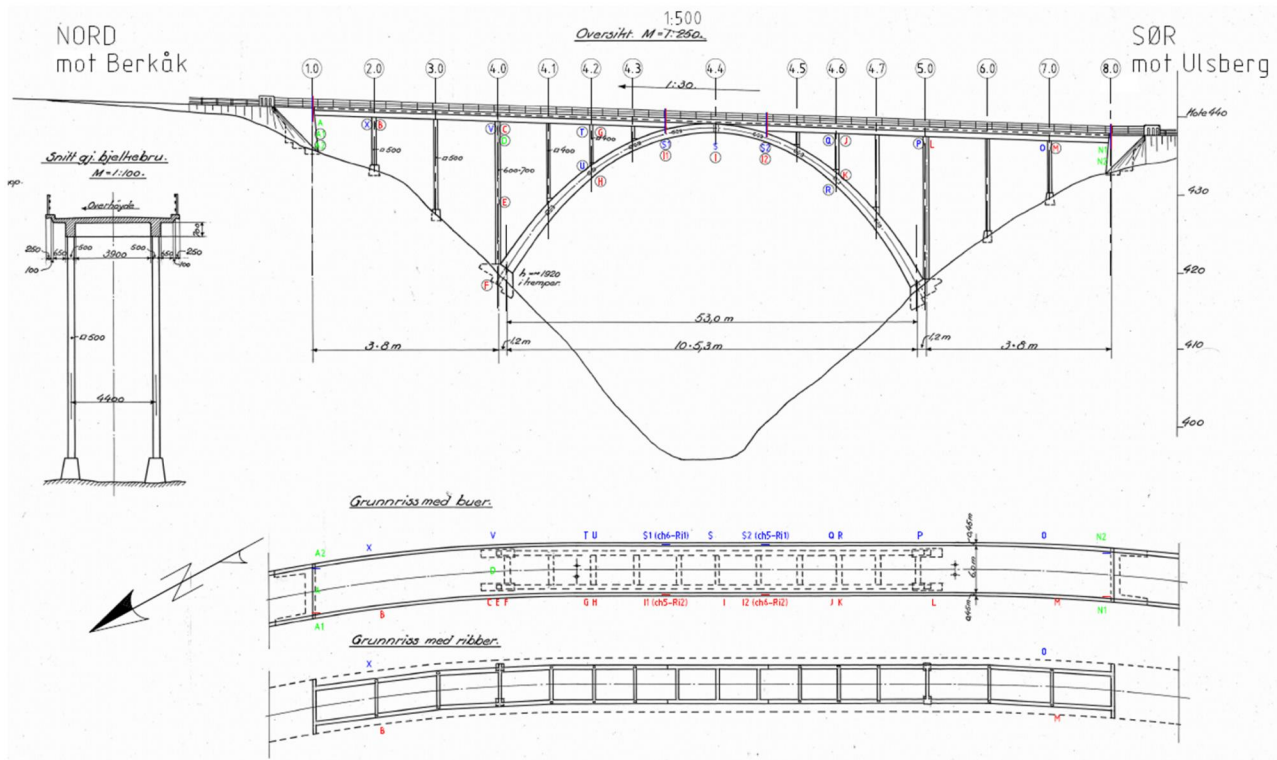

**Figure 6.** Stavå bridge seen from the west and from the top on original drawing from 1942, with location of sensors after 2019 upgrade. Blue on east side, red on west, green distance gauges at abutments. See description in text.

- y-axis: Lateral, positive towards east.
- z-axis: Vertical, positive downwards.

All sensors communicate with a data logger synchronising clocks and polling data. An edge computer located at the north end reads data from the logger, packages it and sends the packages continuously (in scheduled data packages) to a cloud server. The computer also controls a video camera located about 20 meters north of the bridge, taking picture at scheduled intervals.

Temperature and wind data are collected every 10 min, using readings from the closest available location two kilometers to the north, from the Met Weather API by the Norwegian Meteorological Institute (https://api.met.no). As dynamic characteristics change and the structure deforms when the material contracts and expands with temperature, and is also influenced by wind, such information may be important when experts interpret findings. Figure 7 presnts the effect of temperature on the static situation on the bridge, as demonstrated also by Bianchi et al. (2022). Temperature changes will also change the modulus of elasticity in reinforced concrete. Falling temperature implies increased stiffness and will thus also tend to increase the natural frequencies (see for instance, Yanlin Huo, 2022).

As the accelerometers report in their local coordinate system and are not uniformly oriented, all signals must be rotated and 'properly' placed into the global coordinate system. This makes it possible to understand the behaviour in the bridge's longitudinal, transversal and vertical directions, respectively. It also enables calculation of changes in inclination over the global longitudinal and transversal axes. Data feed from the sensors are therefore, upon reception at the cloud, processed to remove noise and bias and to reorient readings correctly into the bridge's global coordinate system, resulting in a transformed and aligned signal of the same resolution (data rate) as the original.

The transformed data stream is then processed in cloud-deployed Python script algorithms running in real time, producing physically meaningful indicator values, accessed *via* the virtual sensors. Lower resolution tailormade, and patented indicators (Hagen et al, 2021) that have been developed and tested in the project, are calculated. One main indicator is termed Energy Response Indicator (ERI). It is in effect identical to the area swept under the curve if the net acceleration signal is plotted in the time domain. When calculated over a second it is identical to the average absolute value of the acceleration over that second.

Another main indicator is termed Energy Characteristics Indicator (ECI), which measures change in the acceleration signal over the period for which it is calculated. It is the area swept under the curve of the derivative of the acceleration signal (the jerk) divided by the ERI, constituting a measure of the average frequency content in the signal, similar to the area centroid in an FFT. Its value ranges from 0 to 2, where 0 is what would be measured during a free fall and 2 when the oscillating frequency is exactly half of the data rate. Also, changes in static angles of inclination over the two horizontal axes are calculated.

ERI and ECI may be calculated on the full acceleration vector or, as used in this paper if nothing else is mentioned, for each one of the three directions. They represent a form of heuristics intended to improve readability, reducing storage needs and allowing for efficient application of predefined alerting rules, or of more advanced ML algorithms. For reasons of transparency a temporal resolution of one second was selected in the Stavå project. However, as these indicators in nature are, or may be made to become, additive, any time resolution would do without affecting the basic motivation. As indicated above, the approach is roughly equivalent to parametrizing an FFT and gives a high-level impression of the development of acceleration amplitudes and frequencies.

### 3.2. A wake-up call

On November 3rd, 2018, the initially installed monitoring system[8] detected a series of crossings where values exceeded previous maxima by an order of magnitude (Figure 8). The amplitude of vertical accelerations in the middle of the bridge span exceeded 0.5 g (5 m/s$^2$) and the calculated vertical displacement during the incident was as much as +/- 4 cm. While no design data was available to quantify the impact this would have on the structural health, it was nevertheless apparent that there was reason for concern. The pattern (speed, amplitude, pattern and pattern similarity, spacing in time) indicated that a column of heavy vehicles had crossed the bridge from the south. The camera was not functioning that period, but the direction could be deduced from the track of the peak of the sensor signals, for instance at the sensor near the north abutment (west on section 1.0) where values gradually built up for then to abruptly vanish. The speed was moderate, around 35–40 km/h, and the individual crossings took place in regular intervals. It was consistent with military transport of heavy equipment. Though never officially confirmed, the incident was likely caused by retreating unit(s) from the NATO exercise Trident Juncture that had taken place in the mountains to the south of the bridge that week (NATO, 2018, 21).

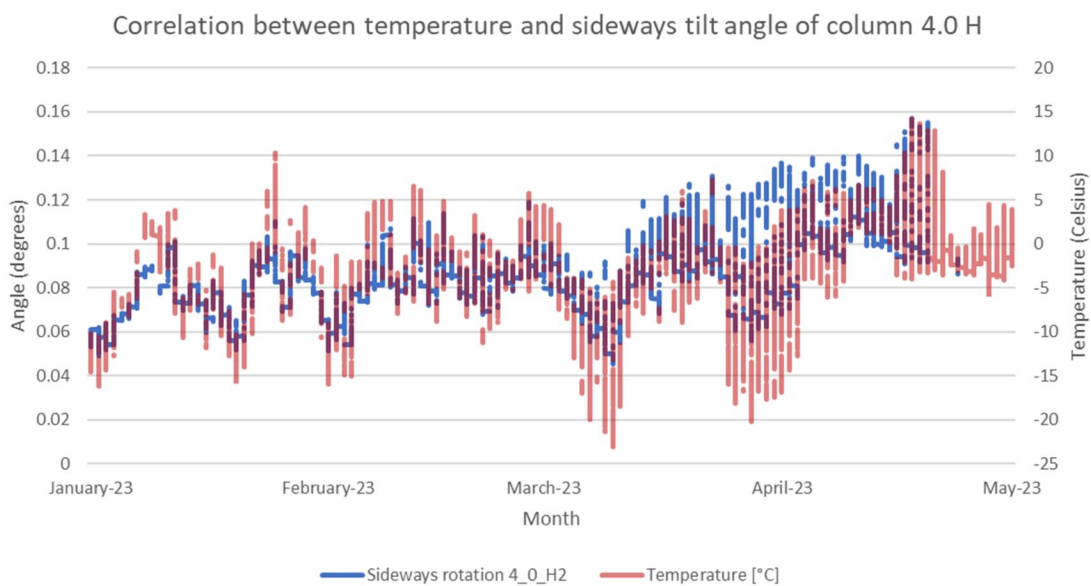

**Figure 7.** Four-month plot, temperature and sideways rotation of sensor E/4_0_H2 (midpoint H2 on column 4.0).

The dynamic responses were so extreme that it was deemed likely that to damages had been caused, and a physical inspection afterwards confirmed the suspicion. A special inspection performed two years earlier, in 2016, revealed one major crack through one of the two main girders, at location I1 on Figure 6 (west side, midway between sections 4.3 and 4.4). The new inspection after the 2018 event revealed three additional, similar cracks, one at location I2 (west, midway between sections 4.4 and 4.5) and two others opposite them, at S1 and S2.

Scrutinizing the events in the aftermath shed more light on the incident. Comparing weekly plots of normalised sum of acceleration amplitudes and frequencies (through ERI and ECI) in the weeks before, during and after the incident showed beyond doubt its significance. Figure 9 shows week-by-week plots of the frequency measure ECI (horizontal) and the amplitude measure ERI (vertical). The observation November 3rd took place in week 44 (October 29th to November 4th).

Studying in more detail extreme values in various frequency ranges disclosed that they 'belonged to' separate crossings. In Figure 10 high resolution (vertical) acceleration time series plots for one high and one low frequency event are superimposed on the low resolution plot for week 44. The embedded 64 Hz, 2 s duration plots show the vertical acceleration of two sensors opposite each other at the top of the arch, mid-bridge. The difference in frequency is clear to the eye—two different vehicles crossing alone, with apparently same weight (as the maximum acceleration amplitudes at the exit are similar), moving at about same speed, only minutes apart, create fundamentally different responses.

That the lowest frequencies were observed late in the column is documented in Figure 11, which shows on the left FFTs of 12 s of each of the two crossings. The frequency shift downward from a dominating 15–20 Hz during the second crossing to a dominating frequency below 5 Hz during the sixth was puzzling. It seemed to point to a (temporary) alteration in the structural behaviour of the bridge. One hypothesis is that the large mass, depending on the path the driver took, for instance small deviations from the centerline when crossing, could create «uplift» and for instance put columns in parts of the bridge in tension, temporarily shifting the natural modes.

The embedded time series of crossing 6 in Figure 10 could support this. The two sensors accelerate with high values in opposite direction, showing that the bridge deck at that section and at that moment twists. The right plots in Figure 11 are FFTs of the four second middle segments of the two crossings, spanning the time when the vehicles would be expected to cross the bridge span. The clearly dominating frequency of ~1 Hz on crossing 6 roughly corresponds to the first lateral natural eigenmode of about 1.4 Hz, adjusted for the added (assumed) mass of the transports. The bridge deck weighs about 300 tons total, roughly half of that being in the main span, while the combined weight of a battle tank and a transporter could reach up to 100 tons. This alone could explain a reduction of natural frequency of the combined system by about 20%.[9]

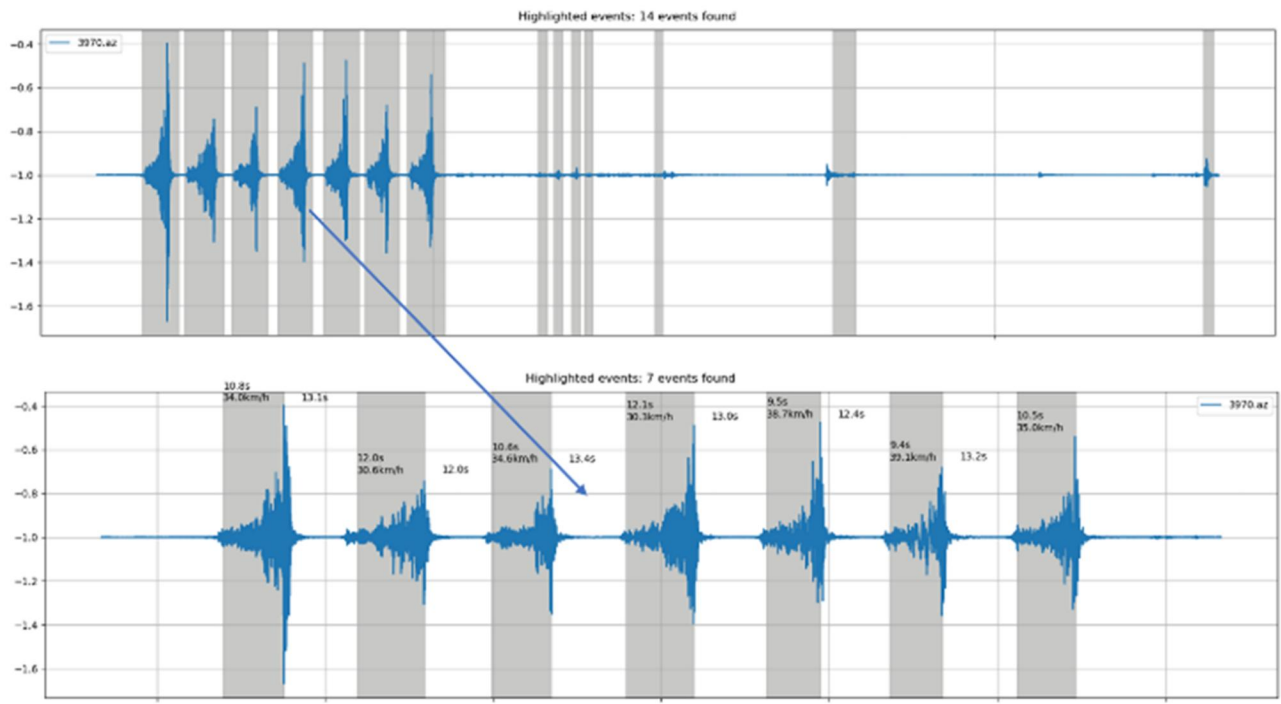

**Figure 8.** Routine machine-assisted screening singled out a set of events, here signal from accelerometer (later moved) at west on section 1. Vertical acceleration in g on y-axis, time on x-axis (time span, seven minutes in bottom detail plot).

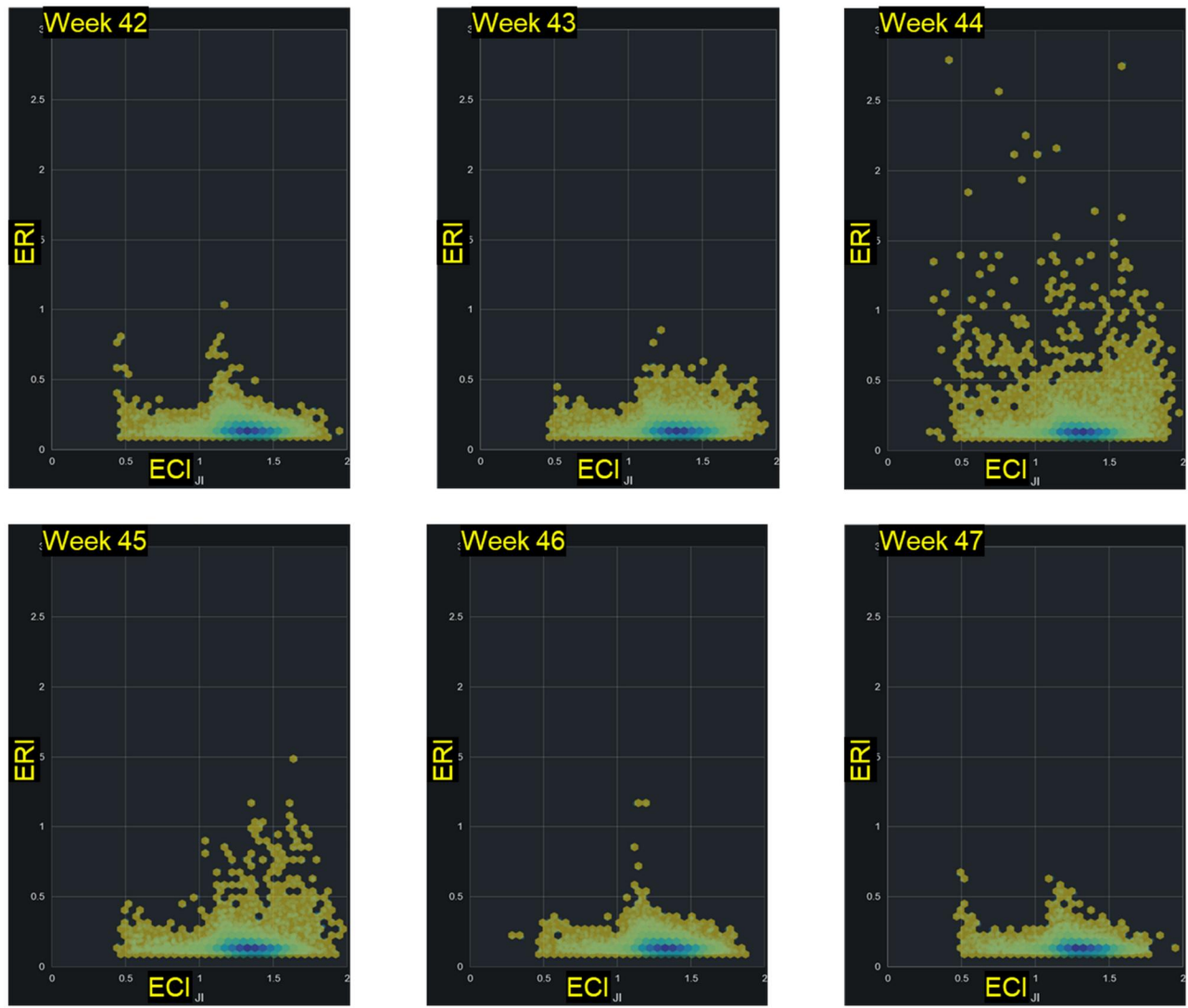

**Figure 9.** Results from sensor S (ref Figure 6), mounted east on section 4.4. Week-by-week scatter plots of ERI (m/s$^2$) on y-axis vs. ECI (dimensionless) on x-axis show all seconds above a noise threshold in the weeks 42–47. Week 44 (upper right) demonstrates a significant overrepresentation of very high values, especially in lower frequencies.

In any case, net acceleration values exceeding 0.5 g combined with frequencies falling towards 1 Hz gave cause for concern. Figure 12 shows details from crossing six, plotting 20 s from the two accelerometers positioned on either side of the bridge midpoint. The bottom graph in each pane shows time series of the accelerations, top graph the resulting calculated translation in millimetres using double integration, upper pane is vertical (z) motion, lower pane lateral (y) motion. This confirms a twisting bridge deck, tilting in phase with sideways motions, amplitude of the vertical displacement of either side close to +/- 4 cm while sideways sway is +/- 1.5 cm. Cracking would be expected from such motion. The vertical and lateral modes are also quite clearly 'coupled', possibly due to some non-linear effects that relate to cracks opening. The frequency also appears to shift downwards at maximum amplitudes, consistent with a heavy load moving, reinforcing the 'added mass' hypothesis.

Returning to a point made earlier about insufficient decision criteria for decision makers, NPRA has a limited basis to set thresholds or acceptance limits for acceleration values, or even deformations due to dynamic, resonant motion. Therefore, even though the engineers involved judged the situation to be serious it was difficult to assess and document the true impact of the event on the structure. There was a mismatch between the metrics from the monitoring system and the metrics familiar to structural engineers.

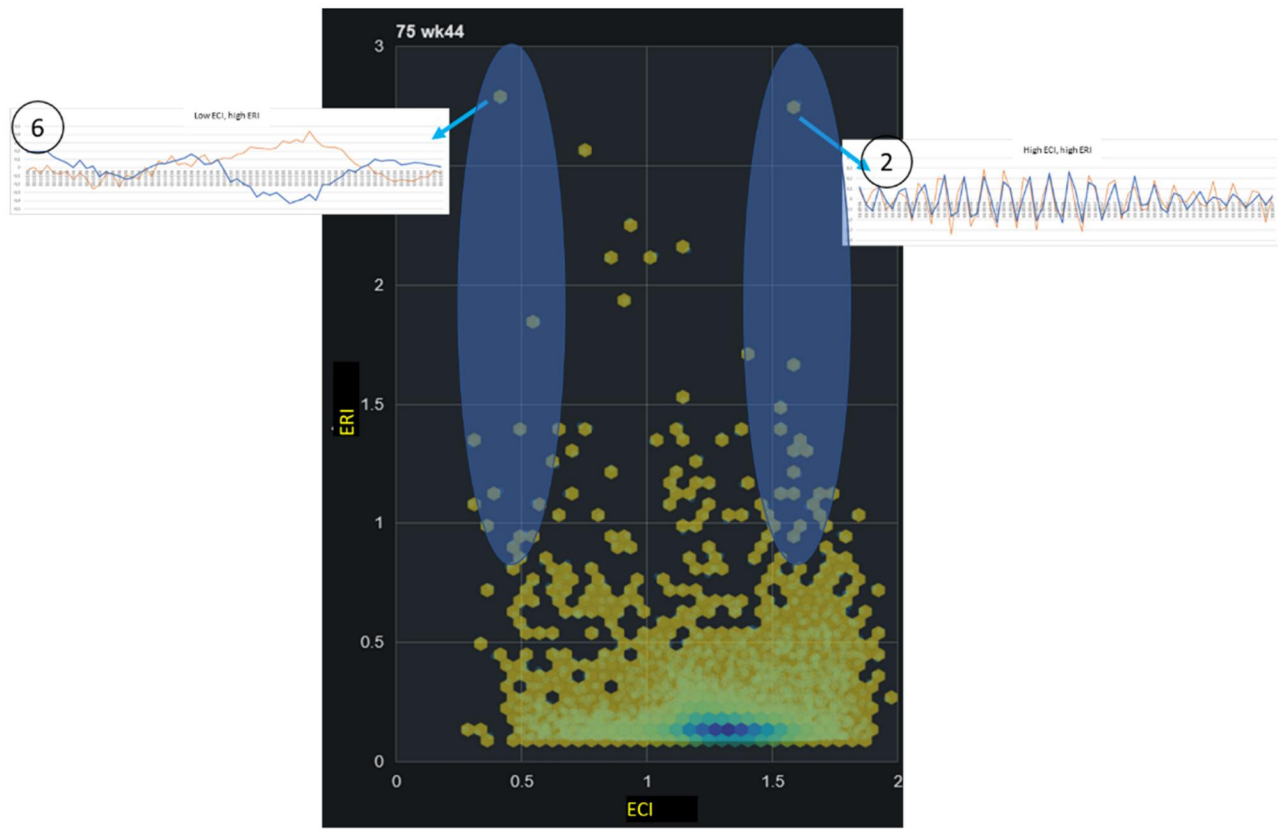

**Figure 10.** Embedded time-domain plot of two seconds in the middle of crossings 2 (11:10:56-57) and 6 (11:12:31-32), from two sensors S and I located on opposite sides of section 4.4 (middle). y-axes show vertical acceleration values in g, scaled +/- 0.4 (2) and +/- 0.5 (6), x-axes show time.

Figure 13 shows result from an admittedly rather crude attempt to close this gap and quantify the effects in more common terms, forcing deformations at sensors I and S in the bridge middle by applying a static torsional moment at section 4.4 and inspecting the resulting stress heatmap.

A report from the technical consultant Rambøll (2016) assumed, based on previous test reports, a yield strength of 20 MPa. At five cm displacement, the von Mises stresses far exceeded this at several locations. The highest concentrations were in areas where new cracks had appeared. The approach is certainly inaccurate, though, as von Mises stress is not fit for this purpose as it primarily applies to ductile materials. The forced displacements were also slightly higher than those calculated using the acceleration time series, and stresses associated with static or quasistatic deformations are not the same as those associated with deformations due to (eigen) dynamics.

Additionally, the model used was idealised, being based on original drawings and using Youngs modules (20 MPa) from the 2016 report and not including major cracks, imperfect boundary conditions and potential plastification in for instance joints. Such effects redistribute loads and relax the structure in statically indeterminate structures like this, see Berger & Haller(2023) for a discussion, and idealised models will tend to exaggerate stresses. But it was nevertheless interesting that, based on the findings, one would expect cracks to appear where they indeed were found, in regions where girders meet arc columns south and north of the bridge's centre.

### 3.3. General usage

Experience from analysing the incident in November 2018 was key when setting out to improve the ability of the system to refine raw sensor data into meaningful information. For acceleration values the following conclusions were reached:

- Raw high resolution acceleration time series to be aligned into the same global coordinate system to produce transformed time series, making it possible to distinguish lateral, longitudinal and vertical eigenmodes, consider rotation over correct axes and to integrate signal into estimated x-y-z displacements.
- The transformed time series from (I) to be subjected to bias removal and processed into low resolution indicators that may reveal dynamic (e.g. amplitudes/ERI, frequencies/ECI) and static (e.g. inclinations) structural features.
- The indicators from (II) to be accessible in the form of output from virtual sensors and subjected to alert rules or ML processing as discussed later, while associated time series from (I) to be stored for potential detailed study.
- Drill-down to be available from the alerts through low resolution indicators (III) into high resolution, stored time series to inspect details of alerted events, seamless switching between time and frequency domains (using FFT) and further investigation through FEA/modal analysis.

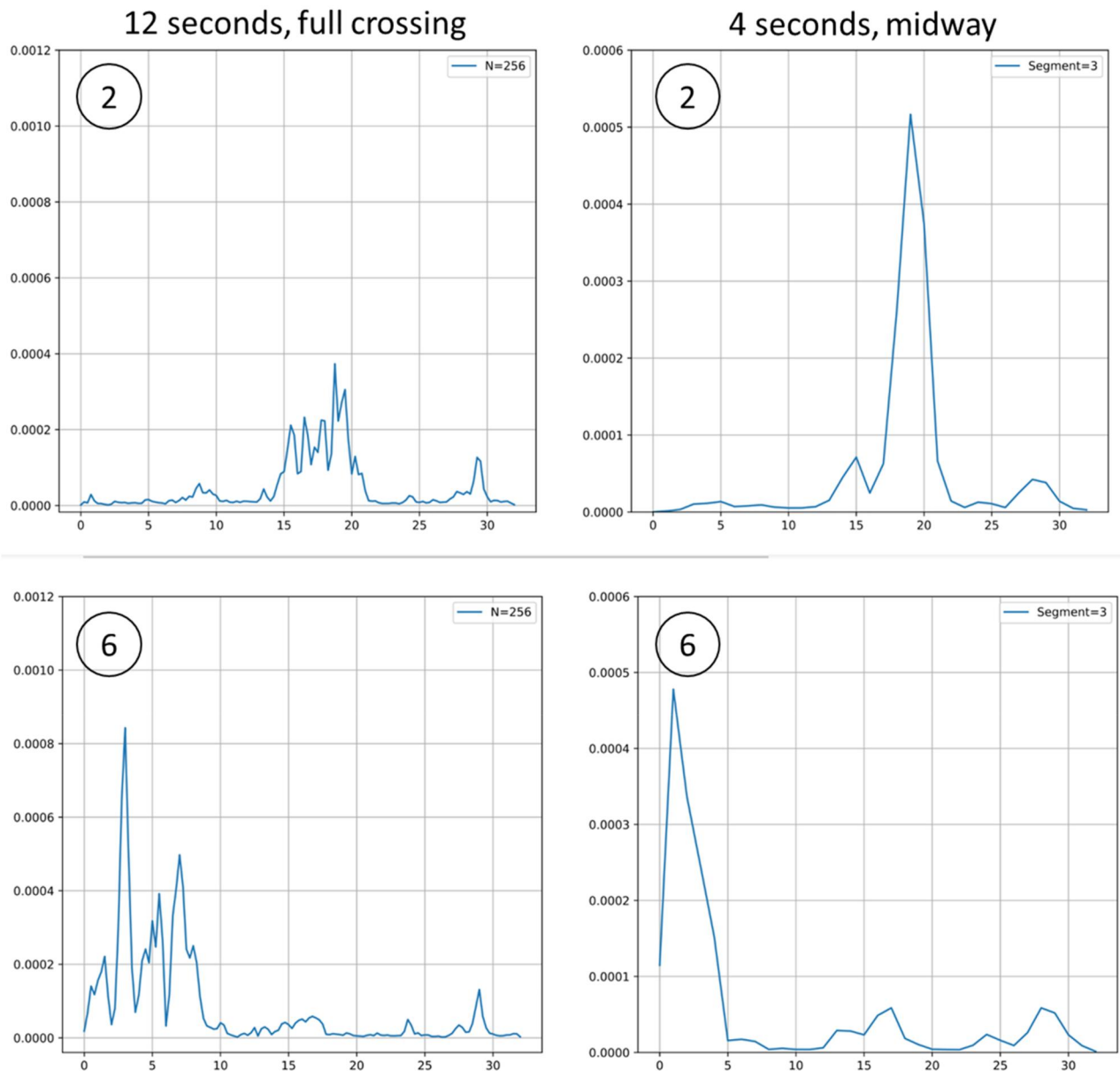

**Figure 11.** FFT (Welch, four seconds overlap) of lateral accelerations on sensor l, top graphs 2nd crossing, bottom graph 6th crossing (see Figure 10). Left graphs show FFT of full 12 s of crossing, right graphs four seconds in the middle of the time series. Vertical axes show a normalised frequency measure, horizontal axes show frequency in Hz.

As common is in diagnostics, correct interpretation of the results relies on understanding how to transform low-level data to high-level meaning. In this case that would mean to understand the underlying physics, and to have some criteria to use when deciding what to do with this understanding. Knowing that an oscillating frequency is generally falling is one thing, but does that knowledge alone give a reason to act? Does the change imply weakening of the structure, or is it simply caused by external factors, for instance environmental or operational?[10] Is the change at all significant, how critical and urgent is it, and when and how should one respond? The overall purpose is of course to support this process of resolution through providing as much insight as possible into the monitored asset, here the bridge. That includes to e.g.:

- Produce warnings according to rules or ML algorithms, flag situations in need of attention.
- Investigate warnings, assess criticality, perform diagnostics, find root cause.
- Gather a history over, e.g. accumulated loads to estimate remaining capacity.
- Show trends to indicate an upcoming situation, enabling preventive action.
- Perform what-if assessments/simulation to predict effects of planned actions.

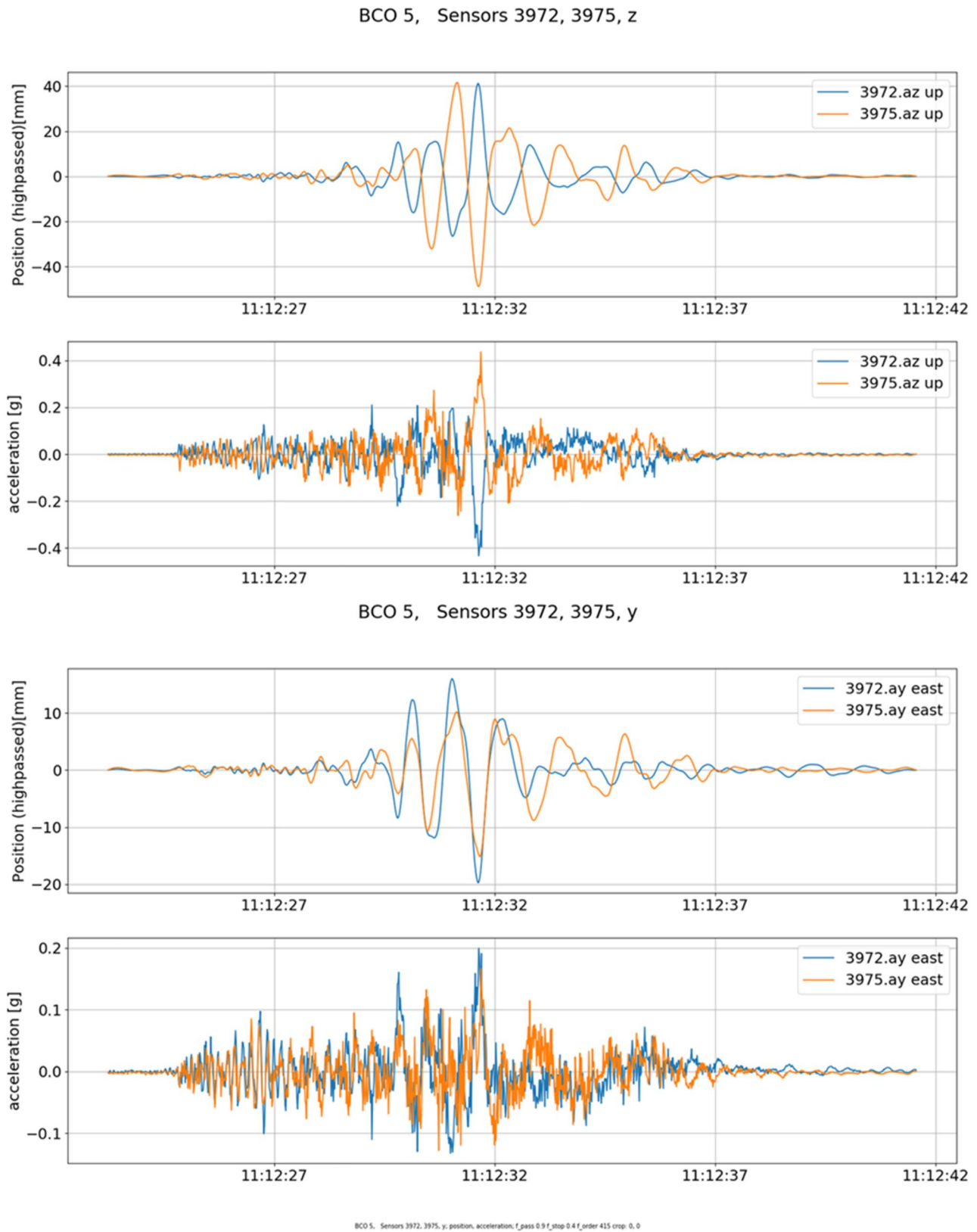

**Figure 12.** 20 s time series of current accelerometer S (in 2018 sensor id 3975) and I (3972) located east and west on section 4.4, upper pane vertical, lower pane lateral (sideways), bottom graph in each pane shows accelerations (in g), top graph deflections (in millimetres) calculated using double integration.

While the above are parts of a sound asset health monitoring, or asset management in general, the focus in the initial stage of the Stavå project was to hold a finger on the pulse on the bridge. The intention was to detect and warn potential anomalies early, with the aim to be more proactive than reactive. This involved threshold-based warnings and trending of selected indicators such as periodic maxima, periodic sums, averages of accumulated impact loads and actions, and average or dominating frequencies. In this process it has proven very important to enable investigations by offering fast and flexible insight into the underlying individual high-resolution signals, whether in time- or frequency-domain, to better understand the behaviour of the bridge in traffic situations. Figure 14 illustrates a typical scenario.

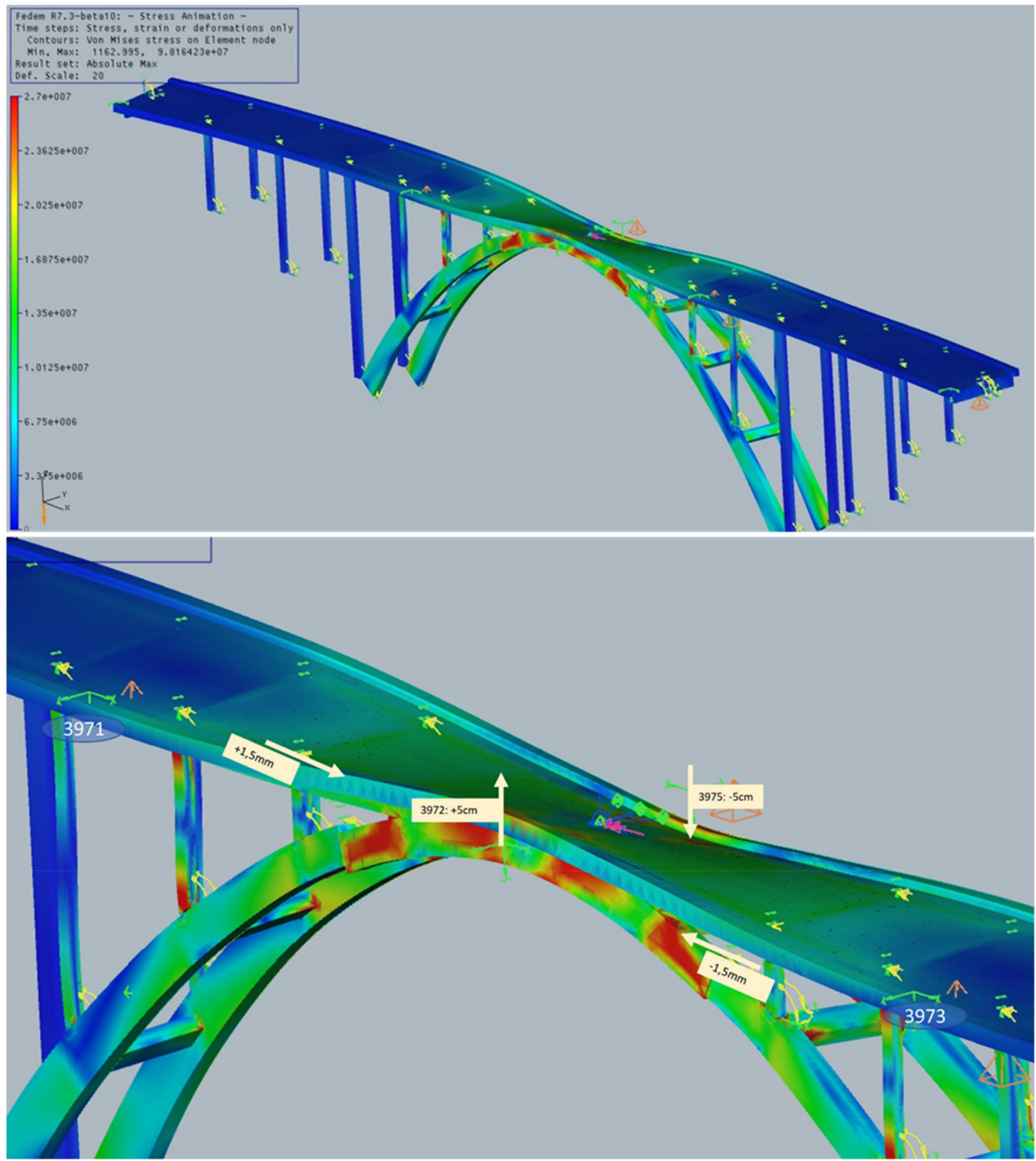

**Figure 13.** Result from applying a static load (torsional moment) on section 4.4 in a FEDEM model of the Stavå bridge, until deformations are similar to those calculated from accelerometer readings. Heatmap scales to 27 MPa.

The panel in Figure 14 is a conglomeration of selected views that could be inspected throughout a virtual investigation, from low resolution to very high resolution:

A. Resolution hours, time span one week. A weekly inspection of maximum directional indicator values (y-axis shows hourly maximum ERI, x-axis shows time) from all sensors mounted on the main span displays the occurrence of some high responses.
B. Resolution hours, time span one week). Selecting the longitudinal view (hourly maximum ERI vs time) reveals that a maximum (below the alert threshold at that time) came during an event on November 23rd, 2021, in the hour between two and three PM, on sensor K/4.6 H2 (section 4.6, west side, arc column).
C. Resolution seconds, time span 40 seconds. After pinpointing the event to 2:09 PM and closing in on sensor K, it becomes clear that the highest ERI value (red, left axis) coincides with a suddenly low ECI value (green, right axis), signifying a (sudden) low frequency event.
D. Resolution milliseconds, time span 25 seconds (64 Hz). A view of the high resolution (64 Hz) vertical acceleration values (m/s$^2$) of K shows that two vehicles likely were involved, that the detected high ERI reading lasts only for a few seconds, and that the response build-up seems to take place almost immediately.
E. Resolution Hz, time span 25 seconds. An FFT of the time series from K, vertical motion, over the same period demonstrates a concentration of energy in the 3.5 Hz band (identical to the first longitudinal natural frequency found in an eigenmode analysis).
F. Resolution milliseconds, time span 25 seconds (64 Hz and 128 Hz on gap gauge). Combining the time series of K in longitudinal direction with the time series from the nearby crack gauge I2 (millimetres on left axis, range 0.25 mm) confirms that the crack opens in the same frequency as and in phase with K, that the pattern is consistent with northbound traffic, and that two vehicles spaced seven to eight seconds were involved.

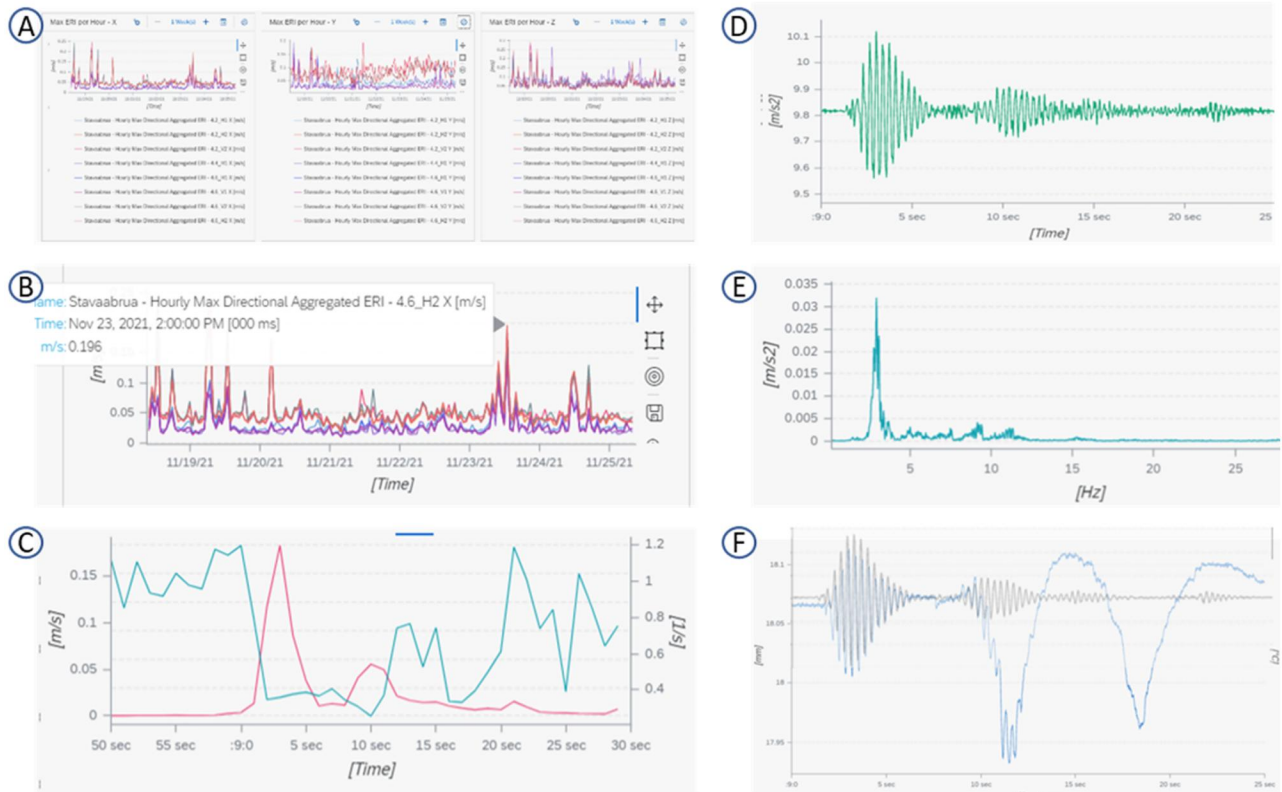

Figure 14. A set of views typically used to analyse a situation. Walkthrough/detailed description in text.

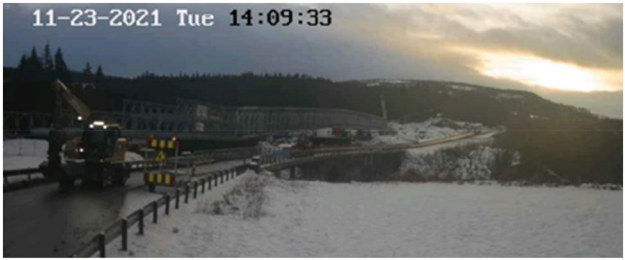

Figure 15. Two excavators crossing bridge cause an impulse on entry, giving rise to large horizontal responses in the first natural frequency, combined with high-frequency disturbance during crossing.

The conclusion was supported by picture documentation. Figure 15 shows a camera capture of two northbound excavators crossing the bridge at the time of the observations. The main hypothesis is that stiff suspension on the excavators combined with relatively high speed caused longitudinal impacts, 'jerks', when they climbed the slope in a ramp, a temporary repair to offload the south abutment, see Figure 16. The impact stirred the bridge into short-lasting eigenmode (first longitudinal) vibration.

## 4. The April incident

### 4.1. Warnings arrive

In normal situations a system notification of exceeded threshold can be statistically expected every now and then, due to the traffic situation, heavy trucks driving fast or in caravan with little distance between the vehicles, special transports with heavy cargo, excavator on belts, mechanical noise from tractors, and even extreme weather. It is simply a statistical fact that unexpected anomalies now and then are just as expected, but in April 2021 the anomalies were physically founded, not statistically. The alerts at the time were based on ERI calculated from the full three-dimensional acceleration (length of the Euclidian acceleration vector). An alert threshold of ERI $0.25\,m/s^2$ was applied equally to all sensors, generating on average about at total one alert every 24 h. A process was ongoing to make individual thresholds based on percentiles for each indicator, but that work was not due to complete until mid-April.

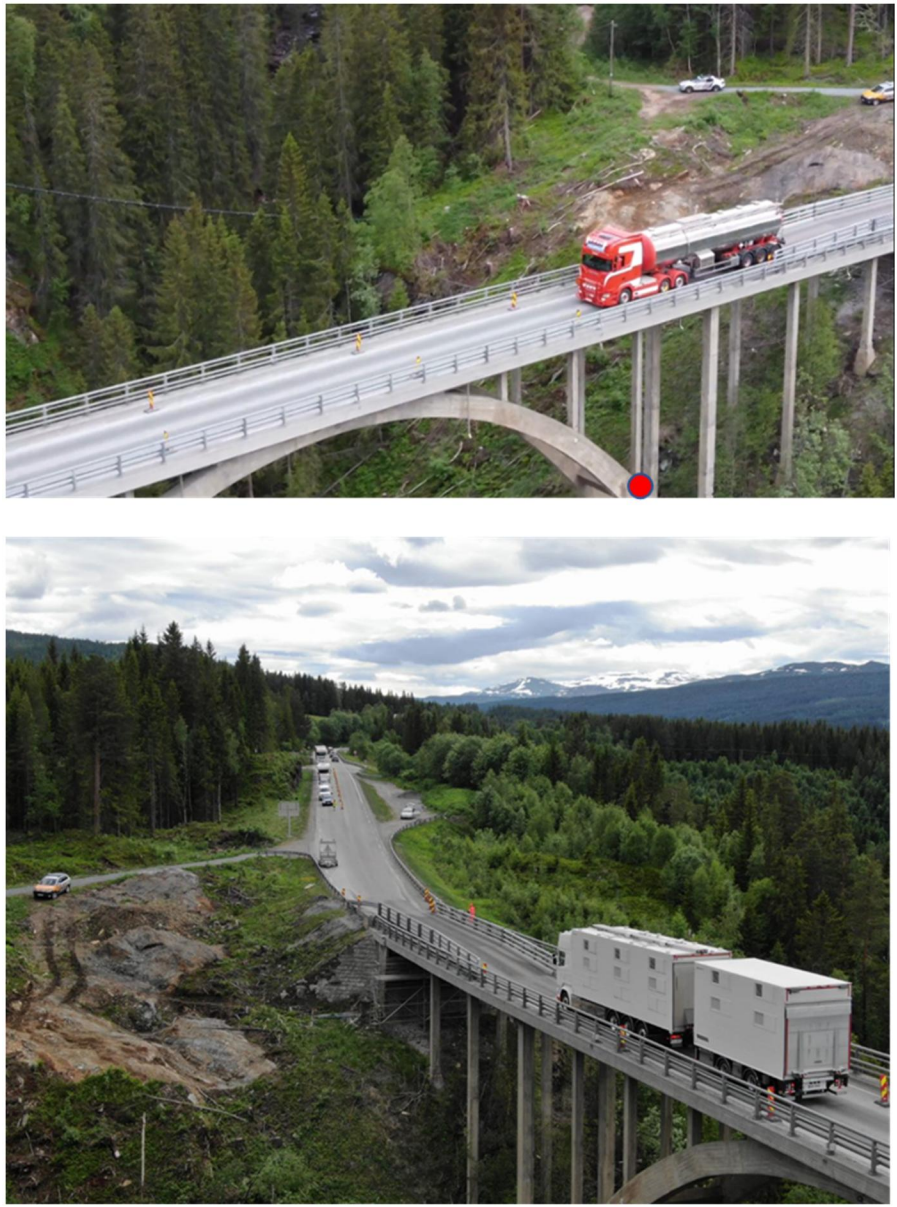

**Figure 16.** Repair done after April 2021, a ramp to offload the south abutment implied an elevation of a stretch of the road, picture showing northbound truck on top (location of sensor K marked by red circle), southbound on bottom. Photos: Kjetil Sletten.

On April 7th, the period between notifications started to decrease in a rapidly developing fashion, mean time between threshold-exceeding events dropping fast. By midday April 8th notifications were generated less than 20 min apart. This reduction in 'Mean Time Between Alerts' from 24 h to 20 min in less than two days clearly indicated that something significant was happening at the bridge. There was no reason to assume that the traffic pattern had changed fundamentally over that short period, so the observation would mean that quite regular cars in the end stirred the bridge in ways only trucks did before. Camera images served as confirmation. The hour-by-hour increase in hourly maxima, representing the ERI value of the most 'severe' second in each hour, from morning April 6th to morning April 8th also gave evidence to this.

The development in maxima over those two days is shown in Figure 17. Preliminary analyses and system tests verified that the alerts were real, and not due to a technical glitch in the monitoring system. Something was taking place that caused the structure to respond increasingly more violentl to quite normal traffic situations. Sensors located far apart showed the same tendency, proving that whatever was taking place influenced the global structure. It was also clear that the supporting arc columns at that time were affected the most (suffix 2 in Figure 17), especially in the lateral (y) direction, and especially in the north (section 4.2), while the

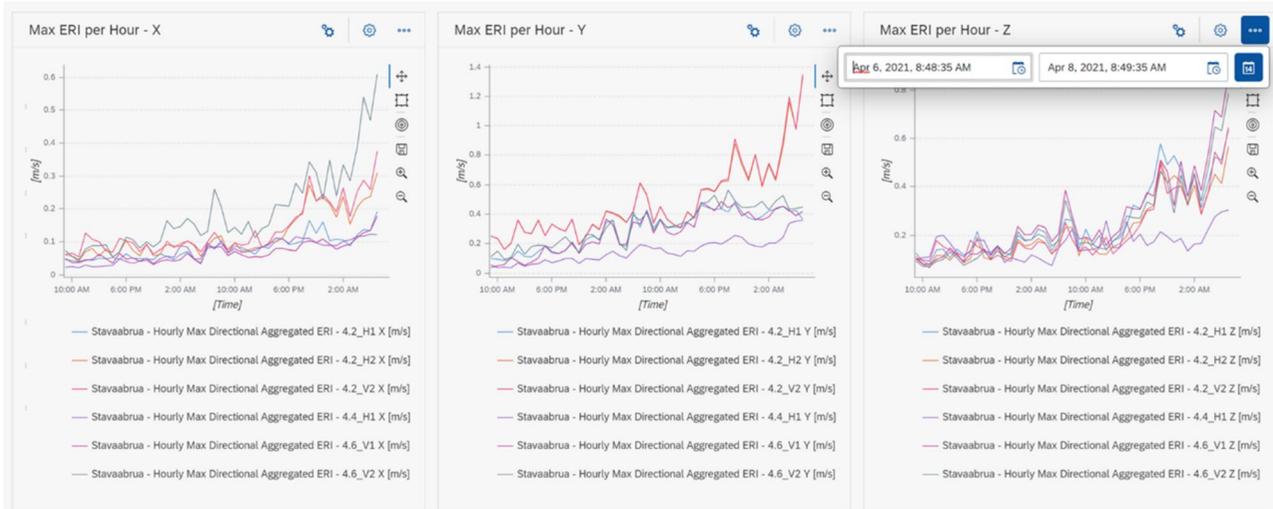

**Figure 17.** Plot of hourly maxima at key positions (from top, G, H, T, I, Q, R, ref Figure 6) on the bridge deck and arced columns in the period April 2nd to April 8th.

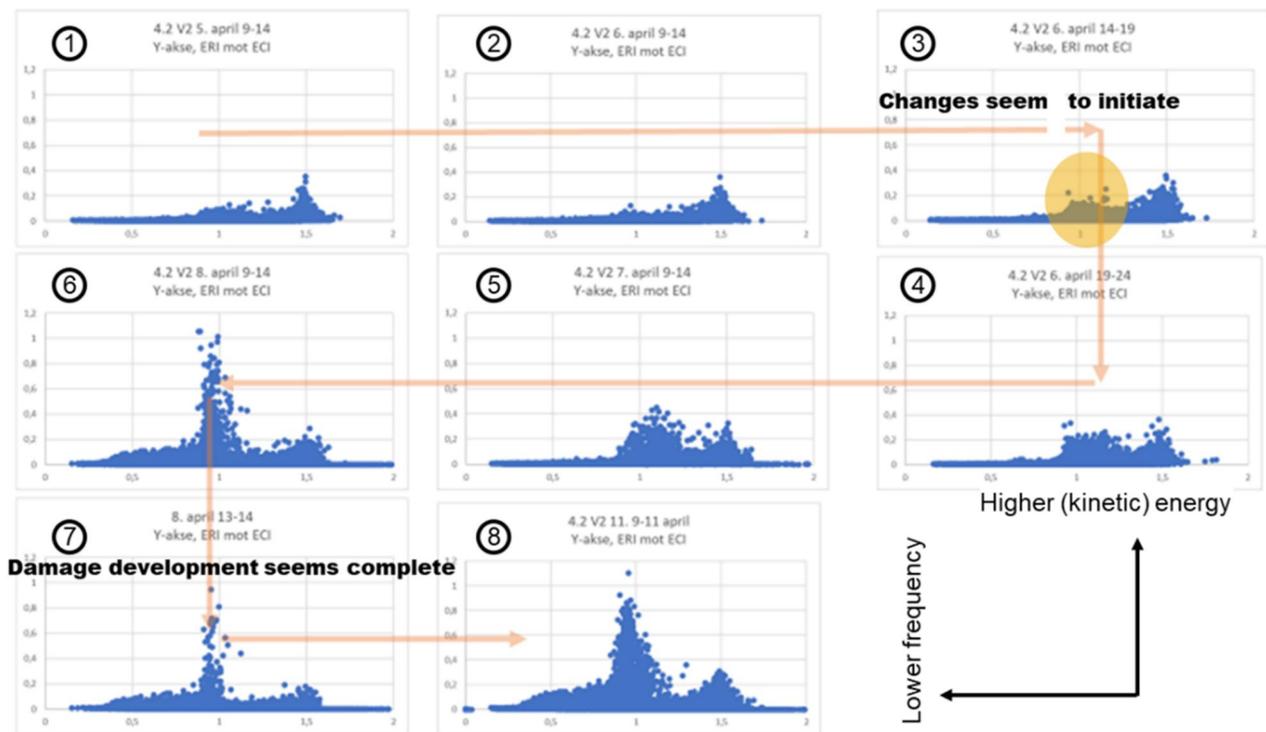

**Figure 18.** From original investigation, tracking development of lateral (y) motion in ERI-ECI on sensor U (section 4.2, east side, arc column) ref Figure 6. ERI (0–1.2) along y-axis, ECI (0–2) along x-axis. Description in text.

bridge deck level initially was far less affected, at least in absolute terms.[11]

Further analyses indicated that the cause of the increase could be the introduction of a new degree of freedom, a relaxation of a major boundary condition like foundation, main structural component, or abutment. The effect was a growth of energy in a mode with lower natural frequency than what, until then, had been normal (Figure 18).

Following the arrow in Figure 18, the scatter plots map ERI vs. ECI for all observations (individual seconds) over selected periods from 9 AM on April 5th to 12 PM on April 11th All plots are to the same scale (0–1.2 on ERI along y-axis, 0–2 on ECI along x-axis):

1: April 5th, 9–14.
2–4: April 6h, three consecutive five-hour periods between 9 and 24.
5: April 7th, 9–14.
6: April 8th 9–14.
7: April 8th, 13–14 (last hour of plot 6).
8: April 9th through April 11th, complete 72 hours of observations.

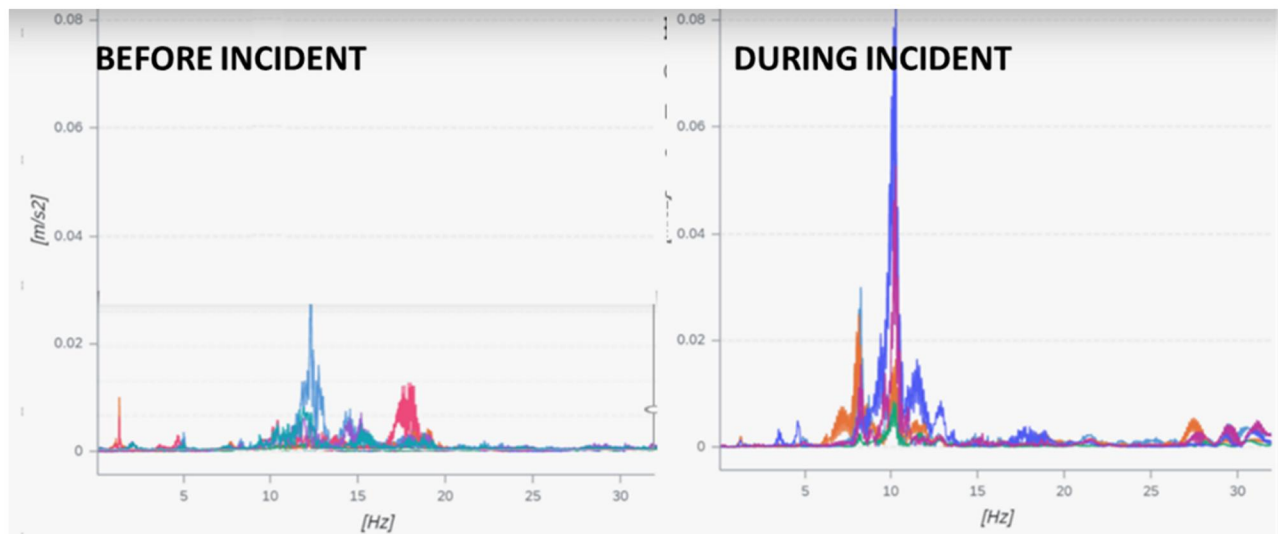

Figure 19. FFT Of the acceleration signal of all sensors during northbound bridge crossings, to the left before and to the right during incident, similar vehicles and speed as identified by pictures. Scales are the same in the two plots, 0.8 m/s$^2$ and 32 Hz.

FFT of selected traffic crossings confirmed the finding, an example of which is shown in Figure 19. While most energy from a crossing before the incident was found in the 12–16 Hz frequency range, the dominating range was now 8–10 Hz. The amplitude in the 'new' frequencies were multiples of those in the 'old'.

### 4.2. Physical inspections

The NPRA responsible acted fast and travelled to the bridge for a physical inspection on the morning, Friday April 9th. At that time the largest responses were registered in the northern half of the bridge, as seen in Figure 17. The inspection therefore concentrated there, but without any findings. It is noteworthy that the bridge deck where the inspector was located was relatively shielded from the significant stirring of the arc column underneath. The situation calmed down a bit in the weekend, as the traffic naturally involved fewer heavy vehicles, and the bridge remained open pending a more detailed investigation on Saturday and Sunday. By Monday morning, these additional analyses had concluded that the largest responses coincided with vehicle entries in the south. Something caused a large global disturbance during northbound traffic that unexpectedly manifested in extreme values particularly north of centre, with sensor H west on section 4.2, about 65 meters north of the south abutment, consistently reporting the highest values.

Close-up studies of the high-resolution time series, such as shown in Figure 20, gave more insights. It turned out that the global response came in a matter of few tenths of a second after first sign of traffic on the bridge. This indicated that northbound vehicles created an extremely high impact loads upon entering the bridge, sending impulses through the structure which in turn excited the 'new' mode(s) seen in Figure 19. Due to the relatively high resolution of the sensors, it was possible to follow the propagation of a 'wave' from the impulse through the structure before the initial impulse died out. As northbound traffic enters the bridge in the eastern lane, the southeast support in the abutment became the chief suspect.

On Monday April 12th there was an extended meeting between NPRA and SAP where the recent findings were presented and discussed, and a new inspection was performed immediately after. Guided by the results from the analyses a serious damage was discovered. The support below the roller bearing at the southeast abutment was gone, the concrete underneath crushed, causing loss of both vertical and sideways support at that location (Figure 21). This had opened new degrees of freedom and 'released' previously contained eigenmodes. As vertical support was lost, wheels of passing vehicles were allowed to act as hammer blows immediately after having passed the expansion joints. These recurring impulse load actions combined with the released constraints caused continual and progressively increasing damage.

The NPRA inspector acted resolutely, closed off the damaged lane with his car, arranged to cordon off the southeast entry point, ensuring vehicles slowed down and avoided the damaged area, made the necessary arrangements with a traffic watch company and did not leave the scene until a manual 24/7 traffic control was in place late that evening. Since then, only one heavy vehicle was allowed on the bridge at a time and the east lane was closed, in an arrangement that would stay in place until a temporary replacement bridge was completed a little less than a year later. Lack of such action could have resulted in accelerated damage progression, even more severely compromising the global structure through the ever-more-frequent and ever-more-powerful impacts. For instance, it would not be unexpected if the southwest bearing, the only remaining in the south, would be damaged by the increased loads and impacts. It could also be that one or more of the cracks at I1-2 or S1-2 would grow, considering the consequences for instance of the 'wavelike' motion seen in the bridge during the event (Figure 20) and the twisting mode documented in November 2018 (Figures 12 and 13).

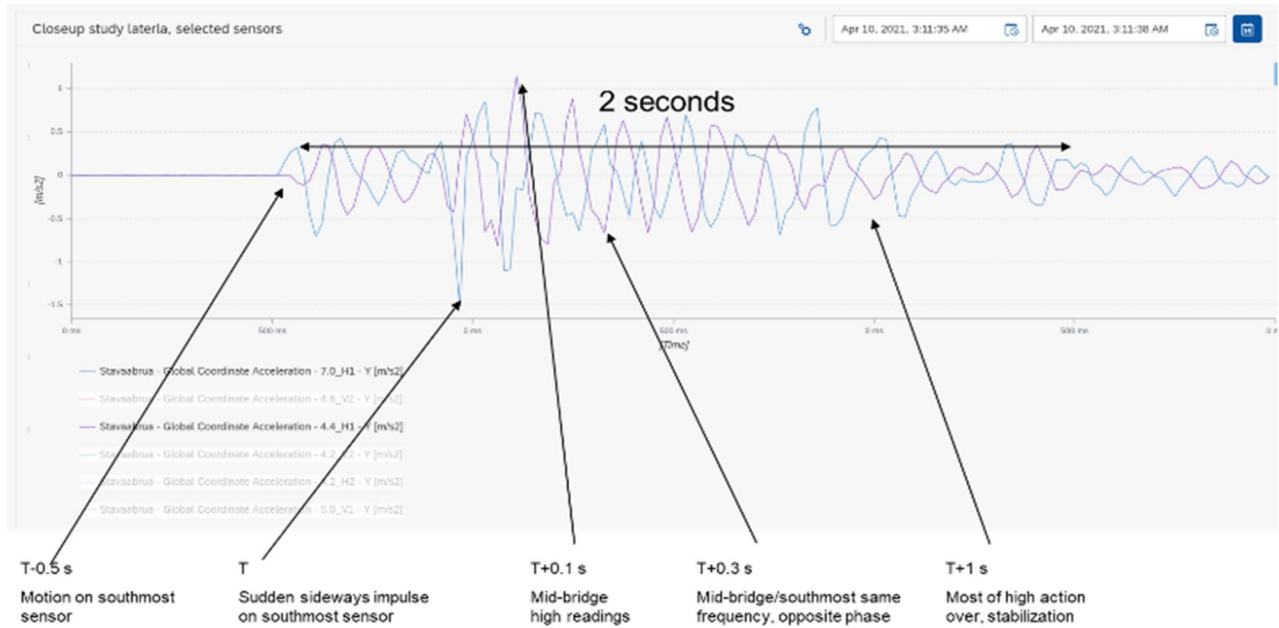

Figure 20. High-resolution view of acceleration signals om two sensors, M mounted at section 7, west side, and I at section 4.4, west side, over a 3-s period. The rapid development demonstrated that an impact occurred immediately after vehicle entry. (acceleration in m/s$^2$ on y axis, scale -1.5 to +1.1.).

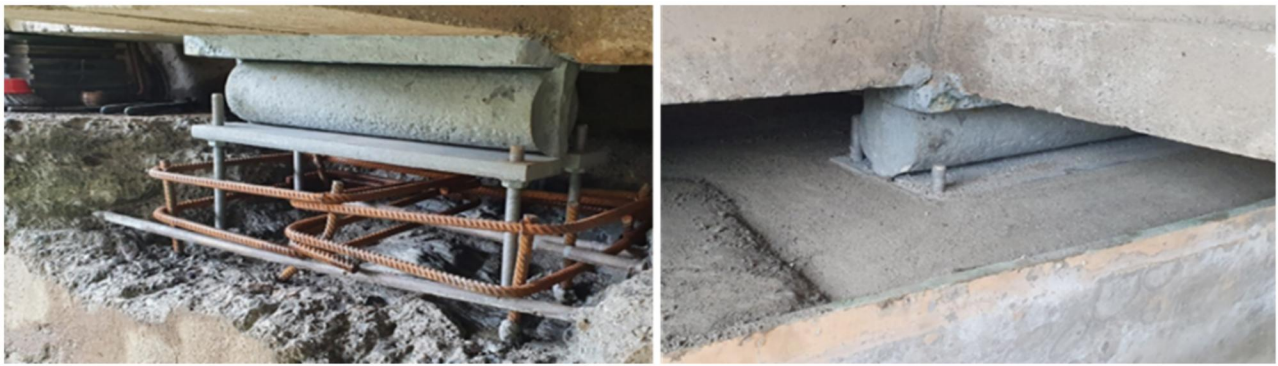

Figure 21. Cause, effect or both? Southeast roller bearing, before and after repair.

## 5. Aftermath, lessons learnt on early detection

A temporary bridge is now in place about 10 metres east of the old bridge, on which there is no more regular traffic. The monitoring campaign has thus changed nature, with no more need for a 24/7 watch. However, after any such event, the question is naturally asked whether the situation could have been avoided, or whether the consequences could have been reduced. A post analysis shows that this indeed might have been possible with full use of available data and models. In lack of relevant metrics and criteria, though, without clearcut and established ways to interpret the findings, this would have implied an elaborate analysis and/or an extended, partially disruptive, and expensive physical inspection, maybe even including surveyors, to be assured that the correct conclusion were drawn before making high-stake decisions. In this case, such a decision could involve preemptively initiating expensive, lengthy, and disruptive repair, or even the same action which was forced in April: Close the bridge for the heaviest vehicles and mobilize 24/7 traffic guard to enforce a one-vehicle-at-a-time policy while procuring and erecting a new (temporary) bridge. This is a difficult call based on facts that are blurry.

### 5.1. CBM vs real-time monitoring: when to replace and when to maintain

The Stavå bridge incident clearly demonstrated the value of SHM in a CBM context, and indeed sound Asset Management in general. By monitoring the structural condition in a continuous manner, it was possible to detect a damage before it turned catastrophic. The aim, though, is to be able to know as early as possible when to initiate preventive measures. A critical decision is to decide the optimal time to renew, when to stop repairing and start replacing, in this case, when to invest in a new bridge and phase out the old one. Given what is known today, this decision should have been taken years ago. The consequences of not having done so have been reduced transport capacity, delays, and increased costs for the replacement bridge, given the short planning and acquisition period. The added costs, including costs to society, have been calculated in NPRA to far exceed ten million Euros. That fact is prudent to contemplate, not to place blame, but rather to learn and improve.

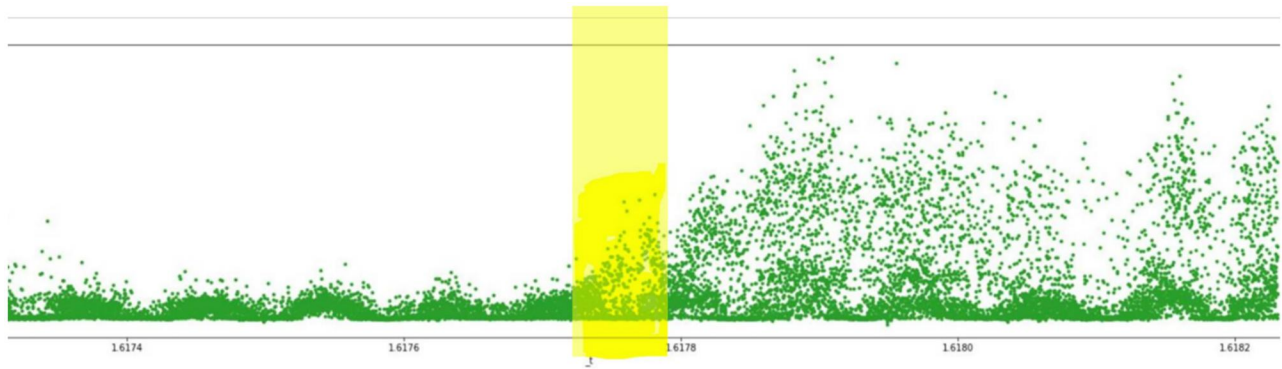

Figure 22. Plot of all individual sideways acceleration values on sensor U at section 4.2, east side, in the period April 2nd to April 12th (timescale epoch time, y-axis scale missing). Yellow field marks period from afternoon April 6th till noon April 7th.

As stated earlier, there are intrinsic uncertainties involved, and trade-offs in need to be made, when deciding whether and when and how to act. One of the largest hindrances to successful CBM and taking preventive measure is arguably lack of understanding of the real situation. Of course, few would resist action if knowing beyond doubt that a critical structure is in the process of being significantly damaged. The reluctance to act is most often due to decision makers having insufficient information of the true asset health, and insufficient information of the consequences of inaction. In addition, organizational risk awareness, and ability to organise according to the actual risk picture, is important. It is not necessarily so that the organisational unit set up to efficiently handle daily operation is identical to that which shall keep an eye on critical assets in degraded conditions, and a crisis management team should not be tasked with doing daily business when situations turn critical.

The norm is therefore that a predictive maintenance strategy, decisions made upon assessed condition and risk, are based on uncertainty. The quality of data, the expected useful life and lifetime models, knowledge of true asset health, and probability and consequences of failure, they are all to varying degrees unknown factors. A professional organization relies on competent personnel able to act before it is too late, using quality software to collect and organise data, and to support decisions. But not least, it relies on a management and company culture able to invest money based on the principles of CBM to avoid excessive firefighting. NPRA is not fully there today, but aims in that direction, organizational changes and training being key. After all, culture tends to eat strategy for breakfast.

### 5.2. Learning through physics

The suspicion arose early in the aftermath of the April incident that there were indications of a possible damage in development at least weeks before. However, at the time the event appeared to come out of the blue, starting early morning April 7th.[12] This was already quite clear from the increased values and steadily more frequent alerts, but the detailed view in Figure 22 of the 10 days leading up to Monday 12th confirmed it. The general and expected pattern with high values during the day and low during the night was broken after midnight between April 6th and April 7th. It was obvious to ask the reason behind such apparently sudden onset.

The search for explanation started alreadyApril 8th. Investigations by NPRA found that there had been construction work involving dynamite blasts on a new road about half a kilometre to the east of the bridge a quarter before one PM on April 6th and right after three PM April 8th. The blasts were detected on the SHM system, as shown Figure 23, but with amplitudes below 0.1 m/s2 they didn't appear significant enough to cause this kind of damage.

Another hypothesis was however formed, related to a trend that was detected beforehand and subject of a separate meeting on April 5th, namely a gradual and slow lean and heel of parts of the bridge. In April 2020, the deployed algorithms were extended to produce average inclinations over both horizontal axes every five minutes. Selected 'calm' periods were used to calibrate angles to 0, in effect setting April 2020 as a baseline. A year after, in April 2021, changes in inclination was seen in most sensors, most notably those placed on the west foundation (F) and arc column (H, I and K), as well as on section 4.0 (E and T). They were consistent with a rotation of the global structure, and changes in the ground/foundation was a potential cause. To investigate this, simulation runs were performed using FEDEM[13] (Norwegian University of Science and Tecnology 2023, SAP 2023b), in which support was removed under the northwest main foundation (Figure 24) at location 4.0 H3, resulting in a sinkage there of about three centimetres.

Figure 25 illustrates the situation after 'transferring' the measured rotations at select locations to a rigid body motion. The result is consistent with leaning and heeling in the direction of one of the main foundations in northwest, which also rotates. Though not conclusive, this supports the hypothesis of sinkage and corresponding change in the static situation (possibly also resulting in crack growth or some other mechanism to 'relax' the structure), in turn having the potential to change the dynamic behaviour of the bridge. The gradual change might have caused a limit to finally be reached. However, the period between calibration and incident was short, not even a full year, and the inclinations change with temperature and frost, so a conclusion at that point would have been uncertain and speculative. This might have been different with a longer measurement period, giving more knowledge about longer term normal behaviour, such as seasonal variations and effects of thaw-freeze cycles.

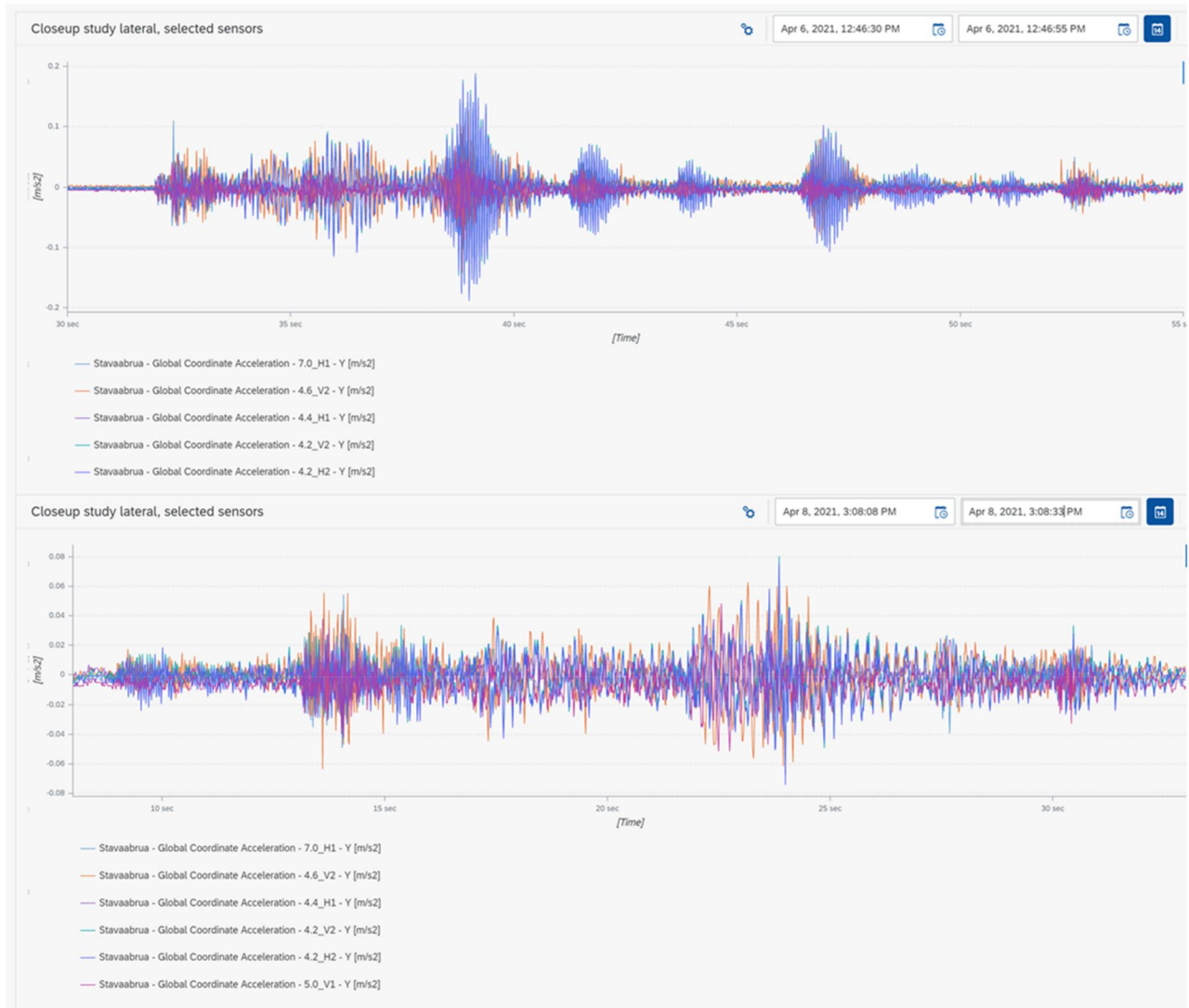

**Figure 23.** Time series (25 s) of lateral accelerations on selected sensors (from the top, M, R, I, T, G and P) during time of dynamite blasts, upper plot from April 6th with visible disturbances from what seems like cars crossing during the event (Stavå bridge being half a kilometre away from the blasts, it was assumed safe and not closed for traffic).

In the aftermath, the hypothesis of foundation sinkage has also been strengthened, at least qualitatively and in theory, by recent findings after bridge closure. Figure 26 shows the maximum values of horizontal plane ERI on selected points over the first six weeks of 2023. Sensor E (4.0 H2) is located at the middle of the column connecting the northwest foundation with the girder supporting the west side of the bridge deck, as shown in Figure 25. Even though the bridge is closed for traffic, this sensor is frequently 'disturbed' and dominates the picture of maximum values with individual horizontal plane acceleration values exceeding 0.1 m/s$^2$, up to ten times higher than any other sensor.

The peaks correlate with strong winds, a significant winter storm coinciding with the maximum values seen on February 6th, making it reasonable to suspect that the cause is wind induced vibrations in the column.[14] This might be expected if the column had lost axial compression, such as would happen in a sufficiently large vertical drop in the foundation, and/or reduced rotational stiffness in the connection with foundation and/or girder. But again, this cannot be confirmed except through physical inspections.

### 5.3. Machine learning

An ever-present dilemma is manifest in monitoring systems tasked with alerting; how to balance false positives against false negatives. The more sensitive the alert limits are, the higher number of false positives is generated, and the higher risk there is that users ultimately start to ignore them, and thus also overlook true positives that require attention. On the other hand, a very strict regime aiming to reduce the occurrence of false positives also increases the likelihood of underreporting true ones. It is a tricky issue of utmost importance. A retrospective analysis was performed, testing how various ML methods could have helped detect anomalies in the period before April 7th, 2021. A data set from January 2021, with all ERI and ECI for the three sensors G (4.2 H2), Q (4.6 V2), and T (4.2 V2), was used as a baseline in a test to discover whether and how the situation in April 2021 could have been detected, validated, and documented earlier. This is admittedly a risky strategy, introducing an inherent problem in applying statistics for such purposes. If a noticeable defect already exists during the 'training' period, it would be built in and normalised, in effect being hidden from view. This is however a separate discussion.

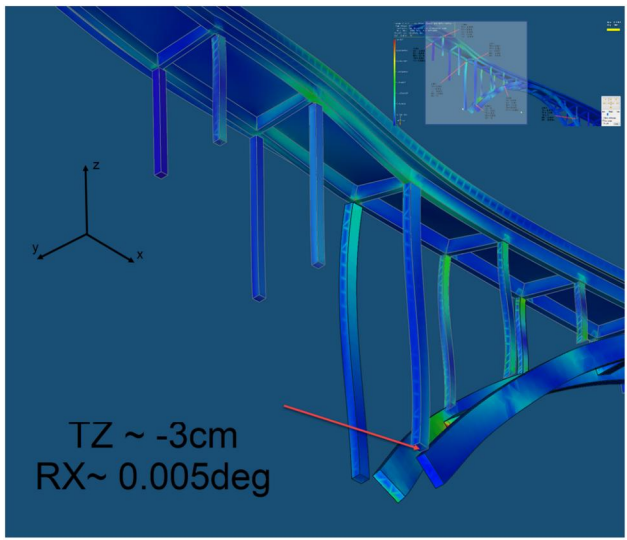

**Figure 24.** Finite element model (using FEDEM) of the Stavå bridge, releasing all degrees of freedom (including vertical support) under the northwest main foundation has the first order effect of a three cm sinkage.

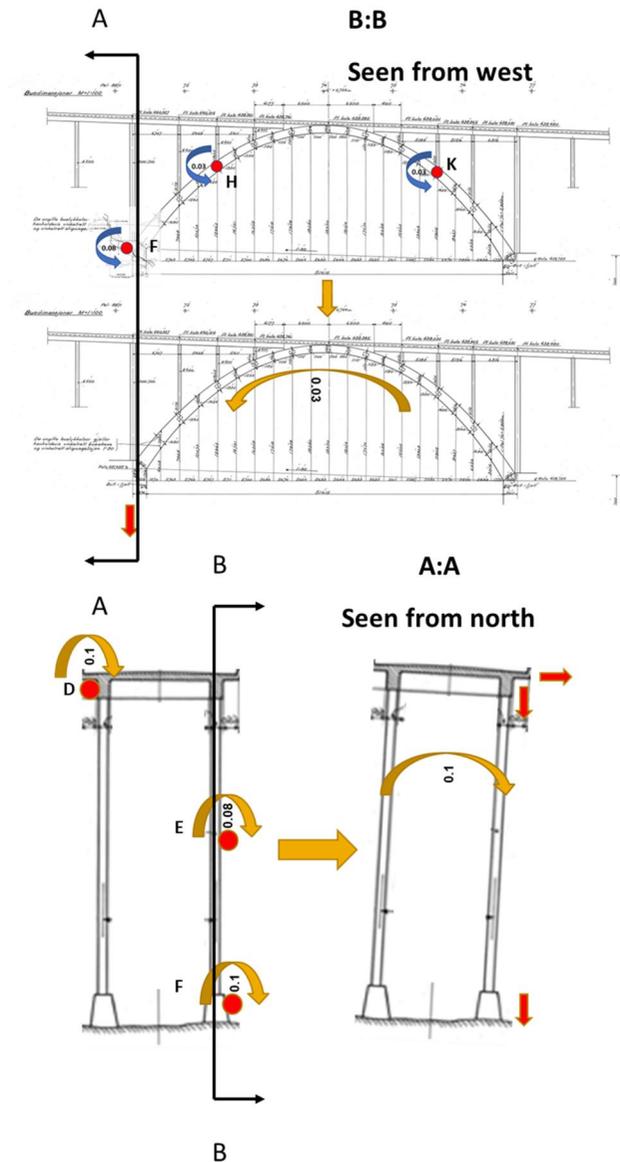

**Figure 25.** Key sensors (lettered references to Figure 6) showed readings consistent with rigid-body rotation of the bridge, counter-clockwise as seen from the west (top), clockwise as seen from the north (bottom).

The first test was quite rudimentary, using the training period to set warning thresholds, indicator by indicator, based on percentiles. The idea was to test the effects on anomaly discovery when using a restrictive alert policy (statistically one monthly alert pr indicator) versus a more liberal policy (one weekly alert). The results are shown in Table 2. If individual thresholds for selected sensors had been based on accepting one monthly warning pr indicator (top), assuming January was a normal month, certain indicators would be seen to have excessive exceedances in March. The indications were however rather weak. On the other hand, accepting one weekly warning pr indicator (bottom) substantially strengthened the proof in March that something was under development.

As seen in the table, while the 'low sensitivity, statistically expected alert once a month pr indicator' strategy produced no warning at all in February, the high sensitivity, once pr week, created two instances of higher-than-expected number of warnings in February, possibly being statistical artefacts. Decision makers might have lost confidence by recurring cries of wolf when no wolf was in sight. In this case, there is no clear objective answer to what is correct—strict or relaxed. It is up to the risk and noise tolerances of the user, and the CBM strategy of the organization to decide, but it needs in any case to be understood that false warnings are a part of life in an organisation committed to catching true ones before it is too late. And it really is too late to sound the warning if the threshold for proof of a true positive is a collapsing structure.

Another test applied a similar approach while using a Manhattan vector (arithmetic sum) of all nine indicators in Table 2. The training period was applied to set a quite 'alert tolerant' 99.99 percentile threshold (mean time between warnings expected to be 10 000 s, about eight pr day). The result is shown in Figure 27 Using this approach, the development could have been clearly seen to start about March 25, with some weak indications some days earlier.

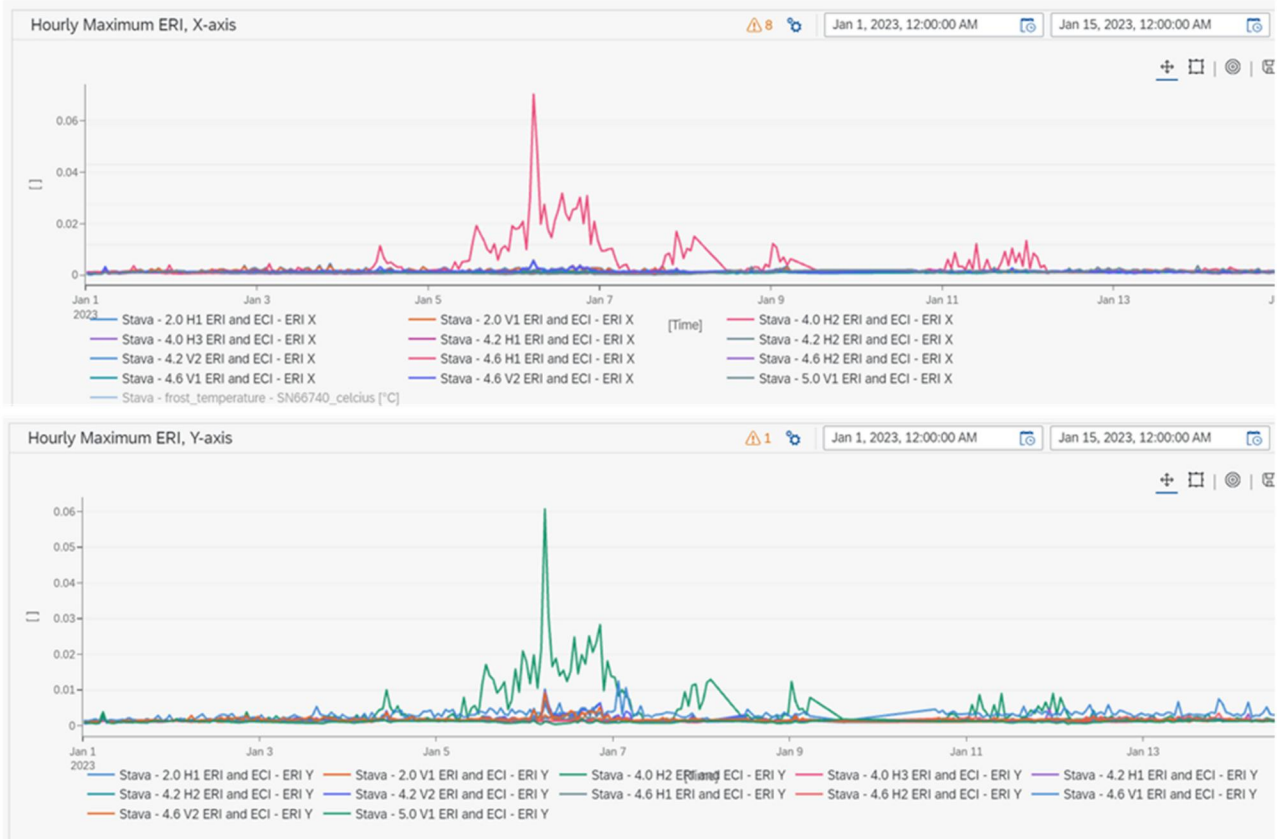

**Figure 26.** Inspecting hourly maxima of ERI in the horizontal plane between January 1st and January 15th, 2023, top being accelerations along x-axis (driving direction), bottom y-axis (across driving direction), peaking values on sensor E (4.0 H2, ref Figure 6) placed at the middle of northwest column.

These approaches are quite simplistic and based on keeping track of a handful of indicators. In situations of slow degradation, where early warnings may come from complex and subtle pattern deviations, this may not be sufficient. To handle a large number of indicators, such as 21 triaxial sensors each with two indicators (ERI and ECI), there is a need for more advanced ML. Another study was therefore performed using the ML algorithm Isolation Forest. In brief terms, Isolation Forest is about using a training period to create an n-dimensional 'shape', n representing the number of features in the model, the shape distinguishing what is defined as within the boundary of normal from that which is defined as outside it. The 'outsiders' are defined as anomalies, each with a score that indicates 'how anomalous' it is.

See for instance Lim (2022) for a general introduction to Isolation Forest and Heigl (2021) for a comprehensive discussion on its use in streaming applications. To illustrate visually, an $n = 2$ Isolation Forest model was built on the feature pair ERI and ECI for sensor Q (4.6 V2) in the y-direction, fifth row in Table 2. The result is shown in Figure 28. Indications around February 28th-March 1st (days 28–29) were found to likely have been caused by temporary issues with this specific sensor, resulting in ECI to get some very low (and anomalous) values. In March (days 52 to 54, March 23rd to 25th) no such issue was found, and the anomalies seem to represent warnings of the structural issue that progressed to ultimately culminate in the second week of April.

This is however only one sensor and one direction, shown here to provide a visual explanation of the approach. It is beyond the scope of this paper to go into details, but for the large scale test an Isolation Forest model was built around 18 features (three sensors and three directions shown in Table 2, as well as the two indicators ERI and ECI). It was also trained on the January 2021 dataset and then run on all observations until end of April. The resulting anomalies were recorded and converted into one common indicator, 'Asset Health'. in this case the daily sum of logarithm of scores as percentage of the January average. Figure 29 shows clearly that something really was in early stages of development around 21st/22nd of March, that it peaked around March 29th and resumed in full April 7th.

While the indicator used and shown in Figure 29 is imperfect to detect degradation of asset health—both technical, operational and environmental factors can play important roles—it is nevertheless an example that demonstrates the value of combining available information into one data set and run analyses on that. However, there are other pitfalls too. Strategies that are best in catching anomalies due to complex interactions also tend to be more difficult in terms of 'transparency, interpretability, and explainability' (Roscher et al., 2020). The more complex the algorithms, and the more dimensions in the underlying data set, the more complicated it normally is to find the exact reason why situations are called to attention as anomalies, and the more important it is to have access to 'interpretation tools [...] to explain the outcome by means of an interpretable representation of the input' (page 42209).

Table 1. Shows the timeline of main events.

| | |
|---|---|
| April 6 | First working day after Easter, nothing appears unusual |
| April 7 | Increase in number of alerts, increased attention |
| April 8 | Rate of alerts increase, hourly maxima started to increase, preliminary analyses performed, NPRA notified, red flag raised |
| April 9 | Physical inspection, nothing found, hourly maxima continue to increase, rate of alerts increase |
| April 10–11 | Situation seems to stabilise at a high level, time series analyses, FEA simulations and modal analyses performed, potential cause found and located to southeast end |
| April 12 | Extended meeting, alarm sounded in NPRA, new inspection found serious damage, lane closed, traffic control implemented |
| April 13–14 | Permanent arrangement with manual 24/7 traffic control set up, speed reduction, one vehicle at a time on bridge, weight restrictions with up to five hour rerouting of heavy transport, work with emergency fix initiated |
| April 15 | Emergency repair ramp / bridge from NPRA contingency depot in Oslo in place |
| August 2021 | Ground work for new temporary bridge started |
| March 23, 2022 | New temporary bridge opened for traffic |

Table 1 Timeline of events, April 6th, 2021, to March 23rd, 2022

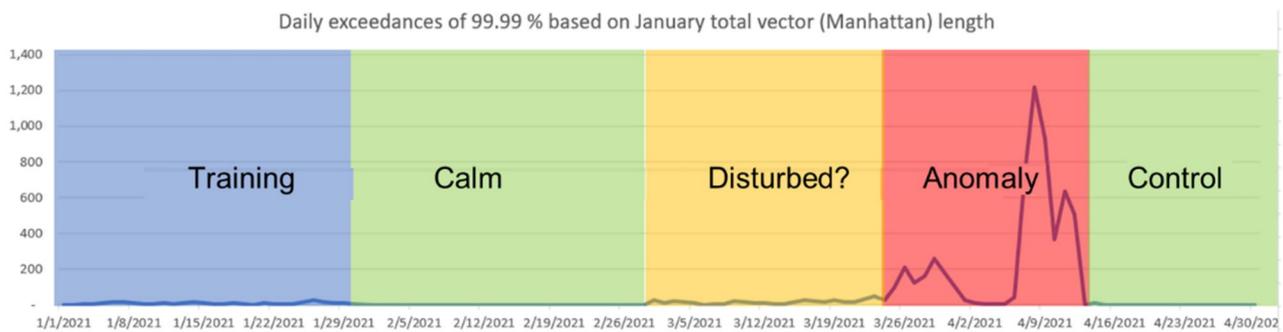

**Figure 27.** Daily alert count with all indicators merged into one Manhattan vector and an eight pr day (99.99 percentile) threshold value for alerts being set in January.

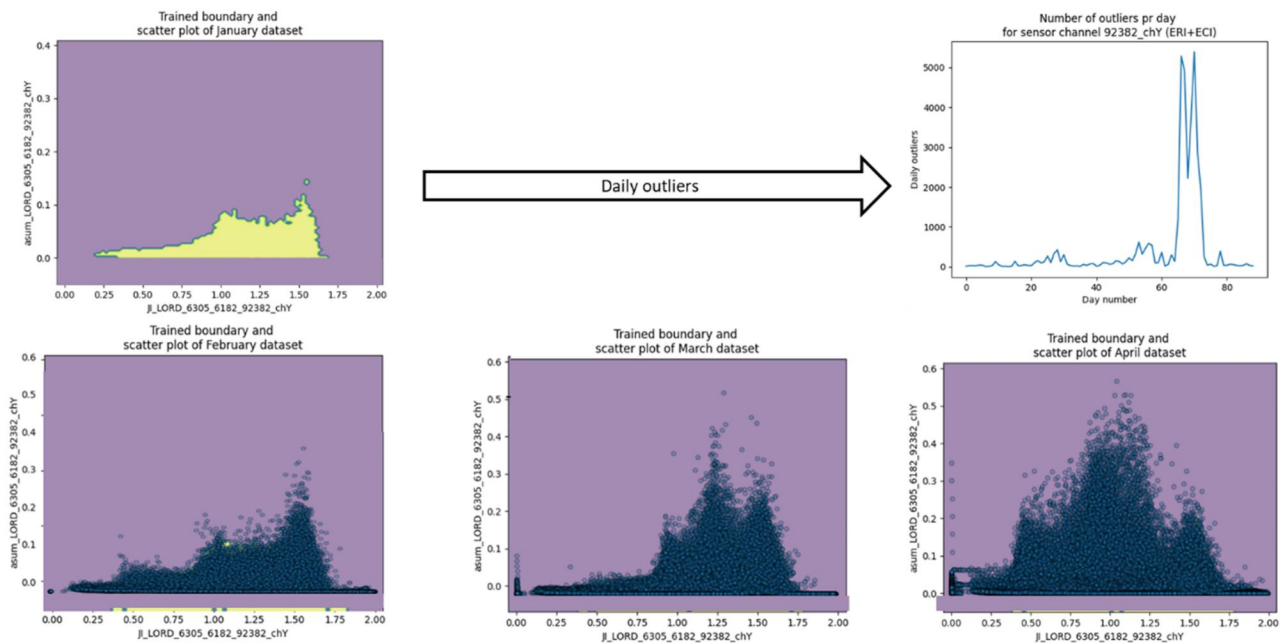

**Figure 28.** Upper left graph, visualized model generated from Isolation Forest algorithm applied to a combination of ERI and ECI for sensor Q, lateral (y) direction, trained on January 2021. Upper right graph shows count of daily reported anomalies, horizontal axis representing number of days since February 1st (through April 30th). The bottom graphs are scatter plots of all seconds, ECI (x-axis) against ERI (y-axis), superimposed on the model for the months of February, March and April.

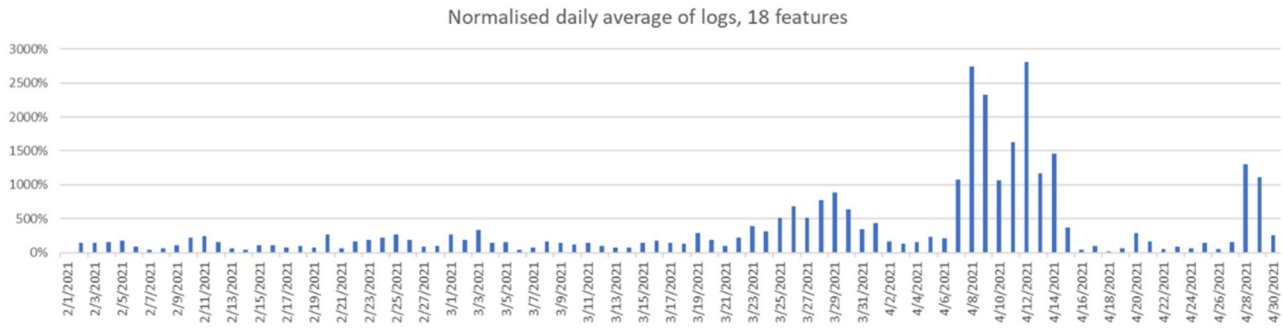

**Figure 29.** Normalised daily sum of logarithm of anomaly scores 'harvested' while running Isolation Forest on 18 features: ERI and ECI for G, Q and T in three directions. The peaks after April 12 (when measures were taken to restrict loads) are related to repair works.

**Table 2.** Three virtual sensors, from top G, Q and T, and three directions. Once-a-month (99.99995 percentile in resolution seconds) indicator-wise alert strategy (top) identifies damage when it is close to fully developed, but March pre-warnings are ambiguous. Once-a-week (99.9998 percentile) warning strategy (bottom) present stronger indications in March, but also some 'noise' in February.

| Sensor | Direction | Alert limit | Jan | Feb | Mar | Apr (before) | Apr (after) |
| --- | --- | --- | --- | --- | --- | --- | --- |
| 4.2 H2 (91) | X | 0.027 | 1 | 0 | 2 | 359 | |
| | Y | 0.218 | 1 | 0 | 7 | 819 | |
| | Z | 0.057 | 1 | 0 | 5 | 651 | |
| 4.6 V2 (82) | X | 0.041 | 1 | 0 | 1 | 1168 | |
| | Y | 0.112 | 1 | 0 | 48 | 409 | |
| | Z | 0.074 | 1 | 0 | 1 | 992 | |
| 4.2 V2 (92) | X | 0.030 | 1 | 0 | 7 | 443 | |
| | Y | 0.214 | 1 | 0 | 8 | 875 | |
| | Z | 0.064 | 1 | 0 | 7 | 603 | |
| | | | 9 | 0 | 86 | 6319 | |

| Sensor | Direction | Alert limit | Jan | Feb | Mar | Apr | |
| --- | --- | --- | --- | --- | --- | --- | --- |
| 4.2 H2 (91) | X | 0.024 | 4 | 0 | 2 | 494 | 5 |
| | Y | 0.177 | 4 | 7 | 19 | 1083 | 0 |
| | Z | 0.052 | 4 | 0 | 5 | 760 | 2 |
| 4.6 V2 (82) | X | 0.031 | 4 | 2 | 7 | 1497 | 3 |
| | Y | 0.090 | 4 | 3 | 91 | 691 | 0 |
| | Z | 0.041 | 4 | 4 | 14 | 1813 | 2 |
| 4.2 V2 (92) | X | 0.023 | 4 | 3 | 14 | 713 | 5 |
| | Y | 0.175 | 4 | 7 | 23 | 1133 | 0 |
| | Z | 0.046 | 4 | 3 | 14 | 967 | 6 |
| | | | 36 | 29 | 189 | 9151 | 23 |

Such interpretation is even more important as diagnostics increasingly rely on what to a large share of users appear as black box ML. Therefore, a combination of deterministic/simulation-based methods and probabilistics/ML is needed. When using advanced machine learning algorithms, for instance Bayesian neural networks, it quickly becomes almost impossible to understand what kind of anomaly is detected. For that, one may need to get into the physics of the situation to disclose which feature(s) of the asset were most involved in causing the anomalous finding, and in what way.

### 5.4. Summary, lessons learnt

Post fact analysis of the Stavå case shows that it could have been possible to detect developing trends earlier, in case of which would have pointed to a need for deeper inspection and maintenance actions that might have prolonged its life. Such early detection of potential failure modes is crucial for good planning upfront for an asset owner like NPRA in the efforts to reduce investment expenses, increase quality, and lower disruption and societal costs. Acting ahead is not easy, though. Among lessons learnt from the Stavå bridge case is that some anomalies are not readily caught by simple threshold exceedance on individual indicators, especially when established criteria to apply during decision making are missing. Early warnings of an adverse development are often subtle changes in patterns of behaviour of several indicators when viewed together. And while one rare observation alone may not be cause for alarm, several otherwise rare observations in a short period of time might.

The experiences from Stavå, both the incident in November 2018 and the one in April 2021, have been important input in the development in SAP of CP as a generic IoT asset monitoring application. When it comes to the SHM system itself, and its practical usage, some of the most important lessons learnt are:

- Continuous (24/7) monitoring and immediate data processing may, even if it seems overly cautious and excessive, be key in detecting rare, early signs of anomalies.
- Removing bias from the sensor itself and correct misalignments introduced during installation is important and will also facilitate continuous tracking of angular changes.

- The use of good low-resolution indicators reduces storage need of high resolution time series, and enables rapid screening of data to improve diagnostics and efficient use of ML.
- Selective, rule-based snapshots of high-resolution time series around key events can be used to pinpoint locations of damages and give input to model-based analyses.
- Switching between time and frequency domains on the same data series, and without delay, is a key tool to assist expert-level diagnostics.
- Replaying events or showing trends with time synchronised data, visualized on 3D models, may prove key to understand readings, especially in complex situations.
- Combining real data with theoretical from advanced simulations (such as FEA) may significantly assist discovery of root causes, e.g. identifying relevant modal shapes.

The human brain has inherent limitations when it comes to how many independent variables can be analysed and compared at the same time, while ML-methods do not (assuming sufficient computational resources). ML can therefore be a very important tool to analyse complex correlations and alert subtle changes, giving early warnings of a future problem. However, a proper balance must still be struck between increasing the likelihood of early detection and reducing the danger of distrusting the system should too many early warnings turn out to be false positives. For this purpose, deterministic methods (direct analysis and physics-based simulation) are important to distinguish critical from non-critical events.

Engineers, that often themselves are the ultimate decision makers in the field of asset management, alternatively advice them, tend to desire an explanation behind results. They often have a need for understanding not fundamentally different from that of scientists, as formulated by Krenn et al. (2022, p. 762). 'Imagine an oracle providing non-trivial predictions that are always true. Although such a hypothetical system would have a significant scientific impact, [they] would not be satisfied'. Detectability is not sufficient alone—explainability is also needed.

## 6. Conclusions

Nothing can completely replace physical inspection of a structure just as nothing can completely replace a physical examination by a medical doctor. Online, real-time monitoring is however equivalent to having a finger on the pulse, 24/7, enabling a deeper diagnostic work to be performed before more elaborate, costly, and invasive physical inspection takes place. It has the potential to improve efficiency and reduce cost, but most of all to increase the likelihood of catching problems in early stages of development. And, as the experience from Stavå shows, registering, storing and virtually inspecting individual bridge crossings, events lasting a dozen or so seconds, may be sufficient to reveal weaknesses in the initial phase of a process of deterioration. The recording of such events may also give clues as to what caused such a process to start in the first place, be that in the present or in retrospective.

For NPRA, the Stavå bridge case demonstrates that the organisation as such has not been (and still is not) as good at avoiding crises as handling them when they occur, at times relying more on individual than collective action. Better separation between doing business as usual and being in (potential) crisis mode is needed to tackle situations like Stavå (as well as Badderen and other similar cases) in the future. Continuous monitoring and analysis, aided by ML, increases the general knowledge of the dynamic and static behaviour of complex structures. This makes it possible for experts to more rapidly and more precisely reach conclusions, freeing up time and effort to be applied to other cases.

This is particularly important in the aftermath of large events, such as floods or earthquakes, where large numbers of urgent tasks may overwhelm a limited number of specialists. As pilot testing of different types of technologies give results, it is important to continue transforming from the traditional reactive firefighting and scheduled maintenance mode to CBM and RBI, integrating real-time asset monitoring into the larger Asset Management system. Such a transformation requires upskilling of staff, since tools and metrics used for interpreting and understanding assets' condition are different compared to the traditional way of executing maintenance. As the medical community gradually embraces telemedicine, so must the Asset Management community gradually embrace 'tele-engineering'.

While the Stavå bridge case demonstrates the value of online, real time and enriched monitoring, it also demonstrates the challenging task of correctly interpreting findings. Further work should focus on ML (and increasingly AI) to enhance the ability to detect and pinpoint structural anomalies, as well as efforts to improve explainability, both through using explainable ML/AI and through integrating automated simulation and (other) deterministic techniques. The aim should be to enable engineers to faster conclude whether or not warnings based on ML/AI anomaly detection are reasons for concern, a particularly important ability in non-linearly behaving systems that change over time, circumstances under which efficient ML algorithms can become very complex and oblique (see for instance Qian et al., 2020; Tang et al., 2020).

This aim can arguably best be resolved through combining transparent system of indicators and notifications, handled in an adapted organization with properly trained users, both humans and machines, and not least a clear strategy and a proactive company culture. It should always be kept in mind that technology is still an enabler, key and increasingly important as it is. People and organizations are still the ultimate decision makers and executers in this field and will be so in the foreseeable future. Technology development and implementation, whether in the field of SHM or otherwise, should not cease to have that backdrop.

## Notes

1. It is beyond the scope of this paper to delve into details on Machine Learning, but suffices it to say here that while ML algorithms may be very good at *detectability*, the ability to identify anomalous situations, they may be similarly poor at *explainability*, the ability to provide to a user a physically meaningful explanation why the situation indeed was classified as anomalous.
2. In the words of Bertola et al. (2020): «An accurate reserve capacity assessment is challenging since the deterministic approach to estimate parameter values, which is suitable at the design stage, is not appropriate to assess existing structures».
3. Later in the paper a 2018 incident, most certainly involving heavy military vehicles, is described. Powers with malign intent could access online monitoring systems to detect and target heavy equipment on the bridge, and otherwise track and analyse their movements.
4. For comprehensive discussions on this subject, please refer to COST (European Cooperation in Science and Technology) 1402 – Quantifying the Value of Structural Health Monitoring, https://www.cost-tu1402.eu/.
5. SAP is a German-based software company, the largest in the world in business applications, of which Asset Management is key.
6. The product is being discontinued in SAP as from end of 2023.
7. Due to the geometry of the bridge, cabling was an expensive solution, and was technically and logistically challenging. Therefore, a wireless alternative was chosen.
8. This first setup was completed in 2016, at that time consisting of 6 accelerometers. After the described incident in 2018 this installation was significantly extended to the 21 accelerometers and 8 distance/crack gauges shown in Figure 6.
9. $f' = f * \sqrt{M/M'} = 1.4\,\text{Hz} * \sqrt{150t/250t} = 1.4\,\text{Hz} * \sqrt{0.6} \approx 1.1\,\text{Hz}$.
10. For instance, had a system for Weigh in Motion (WiM) been in place at Stavå it would have been easier to interpret results, since it then would be possible to associate measured response with (independently) measured load. It would also have contributed towards reducing uncertainty as to the validity and importance of initial findings.
11. This fact, that the largest absolute growth in acceleration amplitudes was seen in the arc, made early-stage detections by inspection on the bridge deck close to impossible. Such early-stage discovery of changes in the bridge dynamics was made possible due to the presence of pre-mounted accelerometers on the supporting arc.
12. The period 25th of March to April 5th of 2021 included Easter, a period with significant leisure traffic but marginal commercial, thus also few heavy transports.
13. Finite Element Dynamics in Elastic Mechanisms, FEA simulation tool developed by the company Fedem Technology. The company was acquired by SAP in 2016.
14. The closest weather station with wind data from that day, Oppdal 30 km to the south, measured gusts with wind speeds approaching 20 m/s.

## Acknowledgment

The authors would like to acknowledge Marit Reiso and Kjetil Sletten for their contribution to the paper Hagen et al. (2022), from which the present paper is extended, for their work on the Stavå monitoring system itself, especially after the November 2018 incident, and not least their roles in handling the situation in April 2021. The authors also appreciate the effort by the many colleagues in both NPRA and SAP, too many to be named, who contributed with development efforts, analytical work, and operational measures. They have ensured that the online monitoring system has been up and running and given important inputs to verify results and diagnose the rapidly escalating situation in April 2021, thus contributing in different ways to ensure that it ended without serious incident. Finally, the authors wish to acknowledge NPRA and SAP for the openness that has made the publication of this paper possible.

## Disclosure statement

No potential conflict of interest was reported by the author(s).